\documentclass[journal]{IEEEtran}

\usepackage[T1]{fontenc}
\usepackage[utf8]{inputenc}
\usepackage{amsmath,amssymb,amsfonts}
\usepackage{nicefrac}
\usepackage{bm}
\usepackage{graphicx}
\usepackage{xcolor}
\usepackage{booktabs}
\usepackage{multirow}
\usepackage{makecell}
\usepackage{multicol}
\usepackage{subcaption}
\usepackage{adjustbox}
\usepackage{algorithm}
\usepackage{algpseudocode}
\usepackage[normalem]{ulem}
\usepackage{url}
\usepackage{microtype}
\usepackage[numbers,sort&compress]{natbib}
\usepackage{tikz}
\usetikzlibrary{positioning, fit, arrows.meta, calc, shapes.geometric, shapes.misc, backgrounds, matrix, decorations.pathreplacing}
\usepackage{pgfplots}
\pgfplotsset{compat=1.18}
\usepackage{mwe}
\usepackage[hidelinks]{hyperref}
\usepackage{doi}
\usepackage{cleveref}
\usepackage{orcidlink}

\newcommand{\matr}[1]{\bm{#1}}

\newcommand{\ms}{\ensuremath{\,\mathrm{m/s}}} %

\makeatletter
\renewcommand\paragraph[1]{\par\smallskip\noindent\textbf{#1}\quad\ignorespaces}
\makeatother

\providecolor{unc-navy}{RGB}{19, 41, 75}
\providecolor{unc-web-carolina-blue}{RGB}{75, 156, 211}

\definecolor{iqblue}{HTML}{1C77C3} %
\definecolor{iqgreen}{HTML}{3FA34D} %
\definecolor{iqamber}{HTML}{E8901B} %
\definecolor{iqyellow}{HTML}{C2A012} %
\definecolor{iqpurple}{HTML}{8E50B0} %
\definecolor{iqred}{HTML}{D6453D} %
\definecolor{iqgray}{HTML}{8A8F98} %
\definecolor{iqteal}{HTML}{1B6E8C} %
\definecolor{iqpink}{HTML}{E377C2} %
\definecolor{iqyellowgreen}{HTML}{9ACD32} %

\pgfplotsset{
 compat=1.18,
 every axis plot post/.append style={mark options={solid}},
 iqaxis/.style={
 width=9cm, height=6cm,
 axis line style={iqgray!75},
 every tick label/.append style={font=\footnotesize, /pgf/number format/assume math mode=true},
 tick label style={color=black!75},
 label style={font=\small, color=black!85},
 title style={font=\small\bfseries, color=unc-navy},
 grid=both,
 major grid style={solid, black!12},
 minor grid style={dotted, black!8},
 tick align=outside,
 legend cell align={left},
 legend style={
 font=\footnotesize, draw=black!25, fill=white,
 at={(1.02,1)}, anchor=north west, /tikz/every even column/.append style={column sep=4pt},
 },
 every axis plot/.append style={very thick, mark size=2pt},
 clip mode=individual,
 },
 iqhbar/.style={
 iqaxis,
 xbar, bar width=7pt,
 y dir=normal,
 enlarge y limits=0.18,
 axis x line*=bottom, axis y line*=left,
 grid=major, ymajorgrids=false,
 nodes near coords, nodes near coords align={horizontal},
 every node near coord/.append style={font=\scriptsize, color=black!75, /pgf/number format/fixed},
 legend style={font=\footnotesize, draw=black!25, fill=white,
 at={(1.02,1)}, anchor=north west},
 },
}

\pgfplotsset{
 iq/winner/.style ={iqred, mark=*}, %
 iq/realssl/.style ={iqgreen, mark=triangle*}, %
 iq/c5sup/.style ={iqred, mark=*, densely dashed}, %
 iq/c1sup/.style ={iqgreen, mark=triangle*, densely dashed},%
 iq/c0sup/.style ={iqpurple, mark=o, densely dashed}, %
 iq/cnn/.style ={iqblue, mark=*, densely dashed}, %
 iq/ssl/.style ={iqblue, mark=*}, %
 iq/random/.style ={iqgray, mark=o, densely dashed}, %
 iq/refline/.style ={iqamber, densely dashed, line width=1pt}, %
 iq/floorline/.style={iqgray, densely dashed, line width=1pt}, %
}

\tikzset{
 iqnode/.style={
 draw=unc-web-carolina-blue, line width=0.7pt, rounded corners=2.5pt,
 fill=unc-navy!4, text=black!88, align=center, font=\scriptsize,
 inner sep=4pt, minimum height=7.5mm,
 },
 iqnodehi/.style={iqnode, fill=iqblue!12, draw=iqblue, line width=1pt}, %
 iqgoal/.style={iqnode, fill=unc-navy!10, draw=unc-navy, line width=1pt,
 font=\scriptsize\bfseries},
 iqnodemuted/.style={iqnode, fill=black!3, draw=iqgray, text=black!70},
 iqarrow/.style={-{Stealth[length=2.2mm]}, line width=0.7pt, draw=unc-navy!72},
 iqarrowdash/.style={iqarrow, densely dashed, draw=iqgray},
 iqedgelbl/.style={font=\tiny\itshape, text=black!55, inner sep=1.5pt},
 iqcap/.style={font=\scriptsize\itshape, text=black!60, align=center},
}

\makeatletter
\renewenvironment{figure}{\@dblfloat{figure}}{\end@dblfloat}
\makeatother
\makeatletter
\if@twocolumn
 \newenvironment{widetable}{\@dblfloat{table}}{\end@dblfloat}
 \newenvironment{narrowfigure}{\@float{figure}}{\end@float}
\else
 \newenvironment{widetable}{\@float{table}}{\end@float}
 \newenvironment{narrowfigure}{\@float{figure}}{\end@float}
\fi
\makeatother
 
\hypersetup{
 pdftitle={IQ-JEPA: A Joint-Embedding Predictive Architecture with a Hermitian Vision Transformer for Sound Speed and Attenuation Estimation from Ultrasound IQ Data},
 pdfsubject={eess.IV, eess.SP, cs.LG},
 pdfauthor={Masashi Sode, Gianmarco Pinton},
 pdfkeywords={ultrasound, self-supervised learning, JEPA, sound speed estimation, complex-valued neural networks, vision transformer, IQ data},
}

\begin{document}

\title{IQ-JEPA: A Joint-Embedding Predictive Architecture\\with a Hermitian Vision Transformer\\for Sound Speed and Attenuation Estimation\\from Ultrasound IQ Data}

\author{Masashi~Sode\,\orcidlink{0000-0002-3685-7378}~and~Gianmarco~Pinton\,\orcidlink{0000-0002-4896-1439}%
 \thanks{This work was supported in part by the National Institutes of Health under Grants R01EB029419 and R01EB036295.}%
 \thanks{M. Sode and G. Pinton are with the Joint Department of Biomedical Engineering, University of North Carolina at Chapel Hill and North Carolina State University, Chapel Hill, NC 27599 USA (e-mail: gia@email.unc.edu).}}

\markboth{}%
{Sode \MakeLowercase{\textit{et al.}}: IQ-JEPA for Sound Speed and Attenuation Estimation from Ultrasound IQ Data}

\maketitle

\bstctlcite{IEEEtran:BSTcontrol}

\begin{abstract}
 The speed of sound in tissue is a prerequisite for well-focused imaging and has diagnostic value, but recovering it from raw pulse-echo channel data is fundamentally a nonlinear inverse problem. Learned solvers are fast yet label hungry. Simulated sound-speed labels are expensive, while abundant real channel data is unlabeled. We propose IQ-JEPA to exploit both data types. An encoder is pretrained without labels to predict the latent representation of masked in-phase and quadrature (IQ) regions from visible context, then fine-tuned on simulated maps. Sound speed appears in the IQ signal as a phase difference, invariant to the constant phase offset. The encoder is a Hermitian vision transformer that operates on the complex signal directly. Its attention is equivariant to that phase and its conjugate-product feed-forward is invariant to it, so the encoder reads a quantity analogous to the one classical coherence methods use. On 79{,}293 Fullwave 2.5 simulations at 2.5\,MHz, pretraining on the 63{,}435 unlabeled acquisitions reaches 15.60\ms{} at 10{,}000 labels. This is a roughly threefold gain in label efficiency over supervised training, growing to over fourfold at 1{,}000 labels. It is about $2.2\times$ below an InversionNet baseline, and 8.71\ms{} at full labels. The gain still grows with more unlabeled pretraining data. Our comparisons point to self-supervision as the dominant factor. The same encoder transfers. Its frozen features expose sound speed and attenuation, and cross-distribution pretraining between layered and abdominal phantoms costs little accuracy. We see this as a first step toward a foundation model for quantitative ultrasound.
\end{abstract}
 
\begin{IEEEkeywords}
 Ultrasound, self-supervised learning, joint-embedding predictive architecture, sound speed estimation, complex-valued neural networks, vision transformer, IQ data.
\end{IEEEkeywords}

\begin{figure*}[t]
 \centering
 \includegraphics[width=\textwidth]{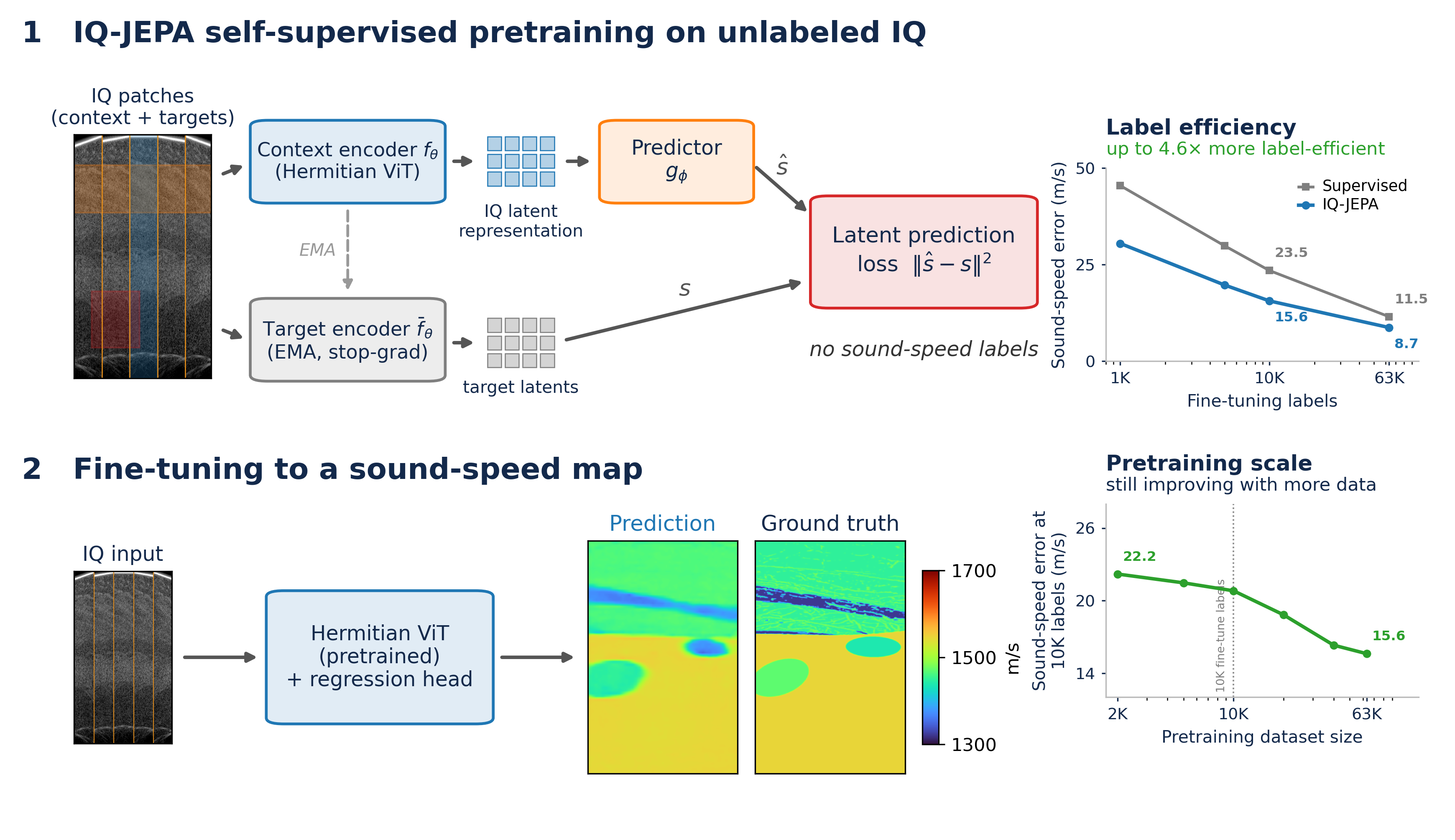}
 \caption{IQ-JEPA at a glance. \emph{Top:} self-supervised pretraining on unlabeled IQ channel data, where a Hermitian ViT context encoder $f_\theta$ and predictor $g_\phi$ match the masked-target latents of an exponential-moving-average target encoder, with no sound-speed labels. \emph{Bottom:} the pretrained encoder, fine-tuned with a regression head to map an IQ acquisition to a sound-speed map. At 10{,}000 labels it reaches 15.60\,m/s versus 23.46\,m/s for matched supervised training, a roughly threefold label-efficiency gain that grows to more than fourfold at the smallest label budget, and the error keeps falling as the unlabeled pretraining set grows.}
 \label{fig:graphical-abstract}
\end{figure*}

\section{Introduction}
\label{sec:introduction}

\paragraph{}
Recovering a hidden physical field from coherent, complex-valued measurements is a learned-inversion problem common to several imaging modalities. Pulse-echo ultrasound \citep{Feigin2024-sos}, seismic full-waveform inversion \citep{Pratt1999-wc, Virieux2009-fwi}, synthetic-aperture radar \citep{DeGuchy2020-ld, Sibler2025-yl}, and magnetic resonance imaging \citep{Rempe2026-kvit} all reconstruct a field from a complex or quadrature signal, and are limited by the availability of ground truth rather than by the availability of raw measurements. In medical ultrasound, the field of interest is the speed of sound, and recovering it from the raw pulse-echo signal is a nonlinear inverse problem.

\paragraph{Sound speed in pulse-echo ultrasound}
Sound speed is the central unknown in pulse-echo ultrasound. Since every beamformer assumes a fixed sound speed value for image reconstruction, actual heterogeneous variation causes an aberration artifact. Sound speed is also a quantitative tissue property that is useful for diagnostics. Therefore, recovering it improves image formation and yields a diagnostic value. As a tissue property, it reflects composition and pathology. In the liver, it decreases as fat content rises, so it tracks hepatic fat and fibrosis \citep{Lin1987-nn, Chen1987-bz, Sehgal1986-aj, Errabolu1987-du}. Pulse-echo estimates on conventional probes have distinguished steatotic from normal liver \citep{Stahli2023-az, Wang2023-hj}. It also carries information beyond the liver, indexing breast-lesion malignancy, muscle composition, and skin \citep{Li2009-breastust, Korta_Martiartu2021-lq, Youssef2018-up}. This matters because the standard alternatives are invasive or costly. Metabolic liver disease is now the most common chronic liver disease, affecting roughly a third of adults worldwide \citep{Younossi2023-wk, Huang2025-mm}, yet its staging still relies on biopsy \citep{Ratziu2005-oq}, and the noninvasive fat reference, magnetic resonance proton-density fat fraction, is not widely available \citep{Zhang2025-nk}. While conventional B-mode ultrasound is widely available, its brightness depends on system and operator settings. Therefore, it is read qualitatively rather than as a calibrated tissue property \citep{Cloutier2021-ze, European-Association-for-the-Study-of-the-Liver-EASL-2024-hl}. Beamforming forms every image by assuming one uniform speed, which real tissue violates. This mismatch degrades resolution and contrast \citep{Anderson2000-soserror, Ali2023-aberrationreview}. Mapping the true speed and correcting the delays measurably improves image quality \citep{Ali2022-aberration}. Attenuation, backscatter, and envelope statistics are further quantitative biomarkers from the same pulse-echo data \citep{Cloutier2021-ze, Oelze2016-xq}, with attenuation likewise tracking hepatic fat.

\paragraph{Prior approaches and the label bottleneck}
Two approaches address this inverse problem. Physics-based methods run iteratively and use no training data. These are full-waveform inversion \citep{Pratt1999-wc, Virieux2009-fwi, Louboutin2017-aw, Guasch2020-cu, Shultzman2022-gr}, coherence-based estimation such as computed ultrasound tomography in echo mode (CUTE) \citep{Jaeger2015-dq, Stahli2020-fn}, and differentiable-beamforming autofocusing \citep{Simson2023-uj, Byra2024-bd}. Learned inverse-mapping networks instead regress the property map directly and run in milliseconds. They range from convolutional encoder-decoders such as InversionNet \citep{Wu2019-inversionnet, Deng2022-qo} to the vision transformer that won the associated Kaggle competition \citep{kaggle-waveform-inversion, Sheoran2025-1stplace}. In ultrasound, learned sound-speed recovery from channel data has been shown \citep{Vishnevskiy2018-va, Feigin2020-yc, Bernhardt2020-wq, Khun-Jush2021-cg, Jush2023-fc, Heller2023-fi}. \citet{Zhuang2023-wc} trained a neural network on beamformed IQ from Fullwave simulations of anatomically derived abdominal models. It resolved adipose, muscle, and liver in silico and produced in vivo sound-speed maps within expected literature values. \citet{Feigin2024-sos} found that the phase of the IQ-demodulated signal effectively decouples operator-dependent effects and stabilizes recovery. This motivates a complex, phase-aware approach. The practical bottleneck for the learned approach is labels, not data. High-fidelity in silico ground truth requires a wave simulation per transmit, in vivo ground truth is largely inaccessible, and labels fragment across transmit schemes, probes, and frequencies. Raw channel data is the opposite. Every clinical acquisition produces it unannotated, so unlabeled channel data is abundant wherever ultrasound is used. Self-supervised learning (SSL) is built for this asymmetry. It pretrains a reusable encoder on abundant unlabeled data and spends scarce labels only on a small readout, as in large language and vision models \citep{Devlin2019-bert, Brown2020-gpt3, He2022-iq}. In the inverse-problem setting, it underlies the scaling results on seismic full-waveform inversion, where foundation models pretrained on large unlabeled corpora transfer with few labels \citep{Deng2022-qo, Feng2024-kg, jin2022unsupervised, Sheng2024-sfm, Liu2024-seislm}.

\begin{figure}[!t]
 \centering
 \includegraphics[width=0.6\linewidth]{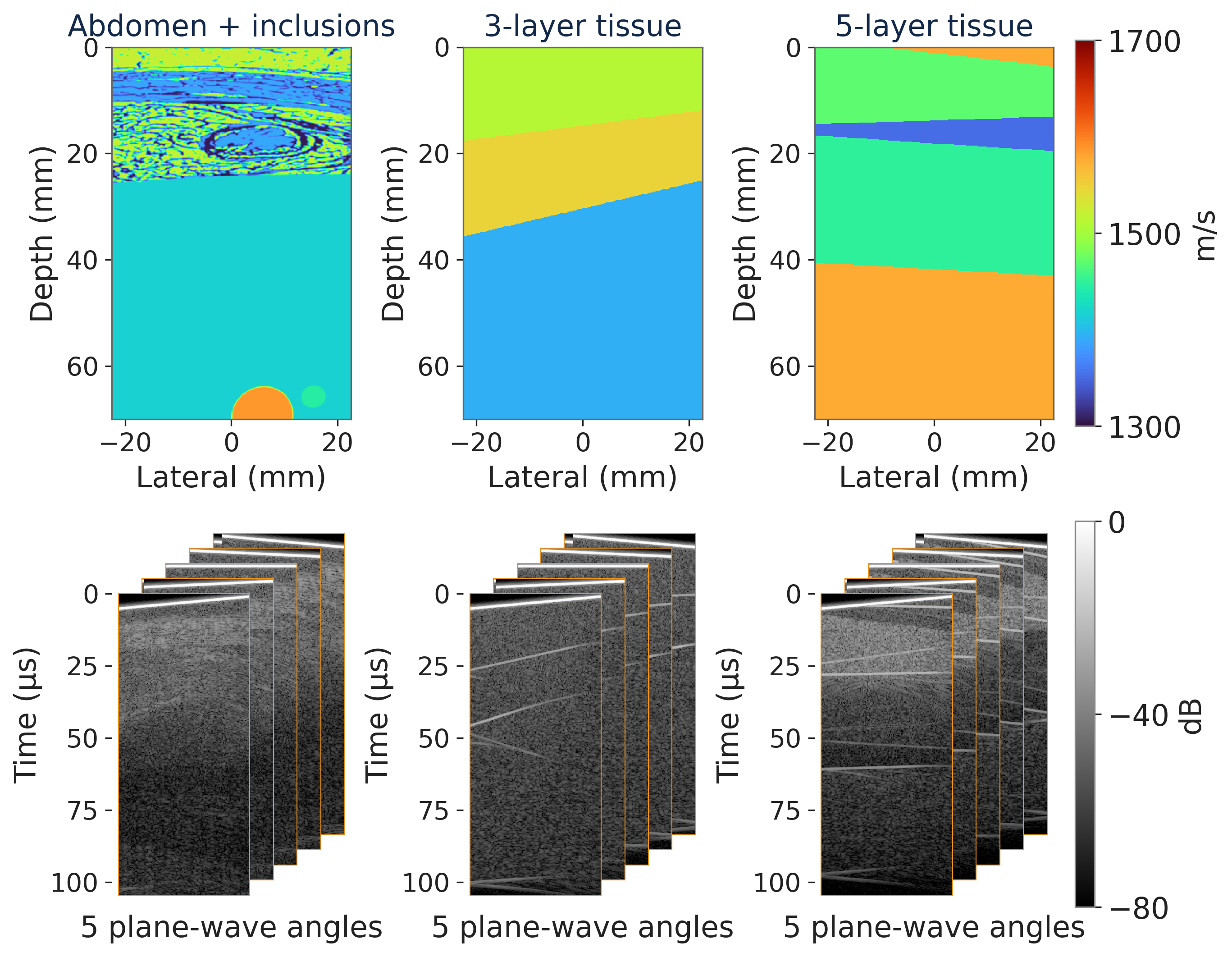}
 \caption{The in silico dataset. Representative sound-speed maps (top) and corresponding IQ channel data (bottom) for the random N-layer and abdominal phantom families.}
 \label{fig:dataset-phantoms}
\end{figure}

\paragraph{Our approach}
We bring this SSL recipe to the ultrasound wave field. We propose \textbf{IQ-JEPA}, a joint-embedding predictive architecture (JEPA) applied directly to complex in-phase and quadrature (IQ) channel data (Figure~\ref{fig:graphical-abstract}). It learns by predicting the latent representation of masked regions from the visible context, with targets from a slowly updated copy of the encoder \citep{Assran2023-ijepa}. Predicting in latent space rather than pixel space lets the encoder discard unpredictable detail and model the structure that matters, which suits speckle-dominated signals, whose exact realization is not predictable from a masked context. Our encoder is a \textbf{Hermitian Vision Transformer}, a vision transformer with a single complex-valued front block and a real-valued backbone, that operates on the complex-valued signal rather than on real-valued radiofrequency (RF) samples. IQ is the natural input because demodulation exposes amplitude and phase as the real and imaginary parts of one complex signal. The phase difference between channels records the travel time that sound speed sets. The encoder is designed around the $U(1)$ symmetry of the demodulation phase, so it preserves the part of the signal in which sound speed lives. What is new in the encoder is the conjugate-product feed-forward. It computes a phase difference that is invariant to the global demodulation phase, analogous to the quantity classical coherence methods use to estimate sound speed. The complex patch embedding, Hermitian attention, and rotary encoding are adopted from the established complex-transformer toolkit \citep{Eilers2023-blocks, Rempe2026-kvit, Sabokpa2026-cvt}.

\paragraph{Contributions}
We evaluate these choices on in-silico Fullwave 2.5 data at 2.5\,MHz. At 10{,}000 labels, self-supervised pretraining lowers sound-speed error from 23.46 to 15.60\ms{}. That is a roughly threefold gain over matched supervised training, growing to more than fourfold at 1{,}000 labels, and about $2.2\times$ lower than an InversionNet baseline (15.60 versus 34.73\ms). The error follows a pretraining-data scaling trend that is still improving at the largest set we tested. The learned representation also transfers. A frozen encoder reads out sound speed far better than a random one, the same encoder fine-tunes to attenuation, and an encoder pretrained on one phantom family keeps most of its accuracy on another. Self-supervision is the dominant factor, seen in the label-efficiency and pretraining-scaling results. Within the encoder, a full $2^5$ ablation ranks phase augmentation, rotary encoding, and the complex patch embedding as the strongest factors. The complex attention and conjugate-product feed-forward are justified on symmetry grounds and matter most when the unlabeled pretraining set is small. Recent ultrasound SSL foundation models, including a joint-embedding predictive architecture, pretrain on beamformed images for anatomical representation \citep{Jiang2025-ultrafedfm, Radhachandran2026-usjepa}. To our knowledge, ours is the first to apply a joint-embedding predictive architecture to the raw complex-valued channel data, before beamforming. Our hypothesis is that this is where the wave physics is most directly expressed. It is also the first to use such an architecture for quantitative sound-speed estimation.

\section{Methods}
\label{sec:methods}

\subsection{Problem setup}
\label{sec:methods:problem}

We estimate a pixel-wise sound-speed map directly from per-element IQ channel data, before any beamforming. Each sample is the raw received echo set from a 5-angle plane-wave transmit sequence at a 2.5\,MHz center frequency. Both choices follow from the simulation budget. A 5-angle plane-wave acquisition needs five transmit simulations per sample, whereas a focused or synthetic-aperture sequence needs 128 or more. Raising the center frequency to 5\,MHz costs roughly four times more computation per sample. This sequence therefore yields the largest number of labeled pairs for a given amount of compute on a 128-element linear array. The target is a sound-speed map $c(\matr{x})$ on a fixed imaging grid, in the range $1300$ to $1700\ms$. All channel data is simulated in silico (Section~\ref{sec:methods:dataset}).

\subsection{Simulated dataset}
\label{sec:methods:dataset}

All channel data is generated with the Fullwave 2.5 finite-difference time-domain solver \citep{Sode2026-to, Sode2026-rd} on a two-dimensional acoustic domain. The transducer is modeled after a 128-element linear array with steering angles $\{-10^\circ, -5^\circ, 0^\circ, +5^\circ, +10^\circ\}$. We synthesize two families of simulated tissue models, or phantoms (Figure~\ref{fig:dataset-phantoms}). The first is random N-layer phantoms with two to six sub-horizontal layers of random thickness, tilt, and tissue properties. The second is abdominal cross-sections from anatomical segmentations of the Visible Human Project \citep{Ackerman1998-xp, Spitzer1996-ly}. These follow the labeled-phantom pipeline of \citet{Zhuang2025-ym} and add random ellipsoidal inclusions. That pipeline segments Visible Human Project cryosections into fat, muscle, connective tissue, and blood at 0.33\,mm isotropic resolution and assigns acoustic properties from the literature, so the simulated medium has an exactly known ground-truth tissue structure. Per-pixel properties are sampled independently per region, not from anatomical priors. The network must therefore learn from wave physics rather than from memorized tissue-property correlations. Sound speed is between $[1300, 1700]\ms$ and density in roughly $[900, 1200]$\,kg/m$^3$. Power-law attenuation follows $\alpha(f, \matr{x}) = \alpha_0(\matr{x})\, f^{y}$ with $\alpha_0 \in [0.1, 0.6]$ and $y = 1$. Sub-resolution scatterers are added to the tissue density distribution. Their echogenic contrast (anechoic, hypoechoic, or hyperechoic) is drawn independently of the material-property inclusions, so a scatterer lesion need not coincide with a sound-speed or attenuation inclusion. For each acquisition, we record the per-element RF channel data, the corresponding demodulated IQ (Section~\ref{sec:methods:iq}), the ground-truth sound-speed map, and a delay-and-sum B-mode reconstruction for visualization only. The full dataset comprises 79{,}293 acquisitions. The test set is a per-phantom-family 10\% draw under a fixed seed independent of the experiment seed, so it is identical across every run and condition (7{,}929 acquisitions). The remaining 90\% is partitioned per experiment seed into training and validation sets (63{,}435 / 7{,}929). Self-supervised pretraining uses only the 63{,}435 training acquisitions, without labels. The validation and test acquisitions are withheld from it entirely, so the encoder never sees the evaluation data in any form. This matters for a self-supervised method, where the evaluation inputs are unlabeled and could otherwise enter pretraining and make the comparison transductive. Fine-tuning draws its labeled subsets from the same training set and selects on the validation set. The test set is reserved for the final evaluation. The validation and test splits are therefore held out from pretraining as well. Sound-speed labels are used only for fine-tuning.

\paragraph{} The solver advances a first-order pressure-velocity system on a heterogeneous medium,
\begin{align}
 \nabla_1\, p + \rho\, \frac{\partial \matr{v}}{\partial t} & = 0, \label{eq:fullwave-momentum} \\
 \nabla_2 \cdot \matr{v} + \kappa\, \frac{\partial p}{\partial t} & = 0, \label{eq:fullwave-continuity}
\end{align}
where $p(\matr{x}, t)$ is the acoustic pressure, $\matr{v}(\matr{x}, t)$ the particle velocity, and $\rho(\matr{x})$ and $\kappa(\matr{x})$ the spatially varying density and compressibility. The local sound speed is then $c = 1/\sqrt{\kappa\rho}$. Frequency-dependent power-law attenuation $\alpha(f) = \alpha_0\, f^{\,y}$ and its causally linked Kramers-Kronig dispersion \citep{Waters2000-bx} are folded into the complex coordinate-stretched derivative operators $\nabla_1$ and $\nabla_2$ through a multiple-relaxation model \citep{Sode2026-rd}. Its relaxation parameters are calibrated per pixel to the target $(\alpha_0, y)$.

\paragraph{} The solver runs on a $70 \times 45$\,mm domain at twelve points per wavelength and a Courant number of 0.4. The reference sound speed is 1540\,m/s. The linear array transmits the 5-angle plane-wave sequence, and receive beamforming uses an f-number of 1.0. Material properties are sampled per region over ranges taken from the measured tissue-property literature \citep{Lin1987-nn, Chen1987-bz, Sehgal1986-aj, Errabolu1987-du, Korta_Martiartu2021-lq, Edwards1988-pp, Youssef2018-up, Zhuang2025-ym}. Sub-resolution scatterers occupy roughly 38\% of the grid. Each ground-truth sound-speed map is rasterized on a $512 \times 256$ grid spanning 5 to 70\,mm in depth. The dataset was generated on 16 NVIDIA L40S GPUs in parallel over about three days, roughly 1{,}150 GPU-hours, and occupies 1.3\,TB on disk. The five N-layer families (two to six layers) contribute 49{,}293 acquisitions, close to 10{,}000 each except 9{,}293 for the six-layer set. The three abdominal families contribute 30{,}000.

\begin{figure}[!t]
 \centering
 \includegraphics[width=0.9\linewidth]{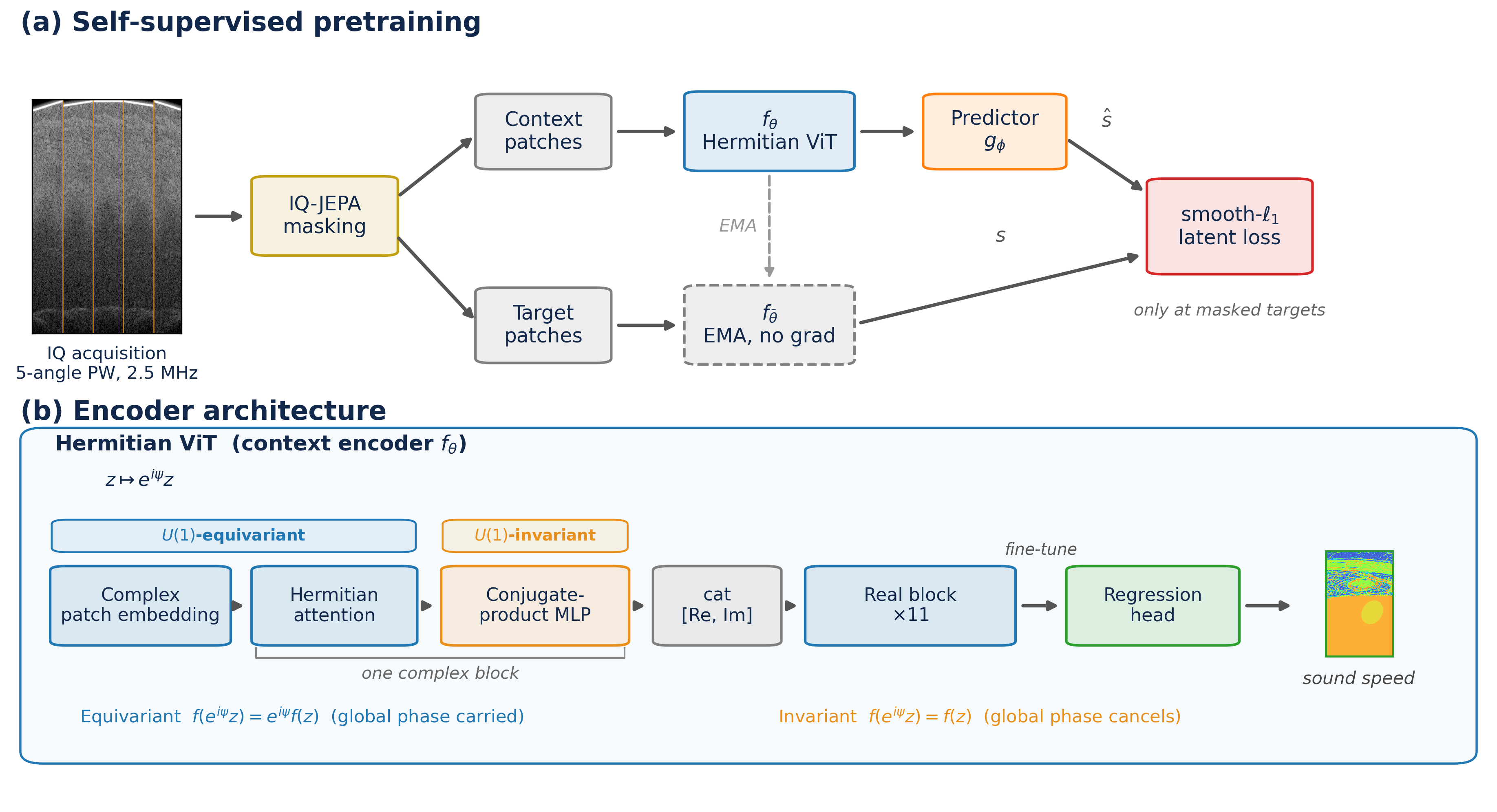}
 \caption{Overview of IQ-JEPA. \emph{Top:} self-supervised joint-embedding predictive pretraining. An unlabeled 5-angle plane-wave IQ acquisition is masked into a context region and several target regions. The context encoder $f_\theta$ (the Hermitian ViT) and predictor $g_\phi$ predict the target latents, matched by a smooth-$\ell_1$ loss to an exponential moving average (EMA) target encoder $f_{\bar\theta}$. \emph{Bottom:} the encoder, a complex patch embedding, one complex block (Hermitian attention and conjugate-product feed-forward), and eleven real-valued transformer blocks. The complex patch embedding and Hermitian attention are $U(1)$-equivariant and the conjugate-product feed-forward is $U(1)$-invariant, the exact structure a phase-invariant sound-speed target requires.}
 \label{fig:method}
\end{figure}

\subsection{IQ representation}
\label{sec:methods:iq}

The received RF signal is demodulated to baseband by mixing with the 2.5\,MHz transmit center frequency and low-pass filtering with a fifth-order Butterworth filter at a 10.7\,MHz sampling rate, without decimation. The filter cutoff is that same 2.5\,MHz center frequency. This yields a complex-valued IQ tensor with explicit in-phase (real) and quadrature (imaginary) channels and 1{,}120 time samples. Skipping decimation keeps the IQ on the RF sampling grid, so the IQ and RF inputs have the same shape and the same token count. The RF-versus-IQ comparison in Table~\ref{tab:label-efficiency} therefore isolates the representation. Demodulation makes amplitude and phase explicit as the modulus and argument of the complex numbers. IQ is a lossless reparameterization of the bandlimited RF and carries the same information. We use it because demodulation makes the phase explicit. The local phase tracks travel time, which depends on sound speed. The complex representation is therefore the natural one for this inverse problem. The phase of the IQ-demodulated signal has been shown to decouple operator-dependent system effects and stabilize learned sound-speed recovery \citep{Feigin2024-sos}. Complex-valued networks are well suited to wave signals, where the phase encodes a time or position difference and the amplitude encodes energy \citep{Lee2022-cvnnsurvey}. The input to the encoder for one acquisition is a single complex image of shape $1{,}120 \times 630$ (depth by the five angles), formed by a fixed spatial reshape. One axis encodes depth (sample time), and the other places the five plane-wave angles side by side along the lateral axis, where each angle is a block of per-element data. The 128 elements of each angle tile the width to 640, cropped to 630 so the $14$-pixel patches divide it evenly. The dropped ten columns are elements 118 to 127 of the fifth angle, and the 1{,}121 time samples are truncated to 1{,}120. Concatenating the acquisition channels into one spatial image in this way, rather than stacking them as separate channels, follows the OpenFWI Kaggle-winning ViT \citep{kaggle-waveform-inversion, Sheoran2025-1stplace}. That solution forced the multiple seismic source channels into a single spatial image for a vision transformer. We adopt three elements of that design, namely this concatenation of channels into the image, rotary position encoding, and the emphasis on dataset size. The reshape preserves two-dimensional locality. The same IQ pipeline is used for pretraining and fine-tuning. Throughout this paper, the IQ panels in the figures display the log-compressed envelope (the modulus $|\mathrm{IQ}|$) for visualization. The network input is the raw complex in-phase and quadrature channels themselves, with no amplitude or phase extraction.

A symmetry follows from demodulation, because the absolute demodulation phase is a free global parameter. Rotating the entire IQ field by a constant phase, $z \mapsto e^{i\phi} z$, does not change the sound-speed map. The encoder is designed so that its operations respect this $U(1)$ symmetry. Each operation is either invariant to the global phase, where sound speed lives, or equivariant to it, carrying the phase forward without scrambling it. This holds for the individual operations. In the simulated data, the global demodulation phase is fixed, so global phase augmentation supplies the variation that exercises the symmetry during training.

\subsection{Hermitian Vision Transformer encoder}
\label{sec:methods:encoder}

The encoder (Figure~\ref{fig:method}, bottom) is a 12-layer vision transformer \citep{Dosovitskiy2020-zl} with a complex front end and a real-valued backbone. We start from the OpenFWI Kaggle-winning ViT \citep{kaggle-waveform-inversion, Deng2022-qo, Sheoran2025-1stplace} and make its early layers complex valued. Every projection in the complex part is a complex-linear layer \citep{Trabelsi2018-complex}. It is a weight-tied Cauchy-Riemann map, with real and imaginary weight parts $W_r$ and $W_i$ acting on the complex input $x = x_r + i x_i$, using half the parameters of an unconstrained $2 \times 2$ real map:
\begin{equation}
 y = (W_r + i W_i)(x_r + i x_i).
 \label{eq:complexlinear}
\end{equation}
The encoder is designed around five components, four learned and one a training-time augmentation.

\paragraph{Complex patch embedding} A strided complex convolution maps the IQ input to a sequence of complex tokens,
\begin{equation}
 z = (K_r + i K_i) \ast (x_r + i x_i),
\end{equation}
with real and imaginary convolution kernels $K_r$ and $K_i$, a $14 \times 14$ patch that tiles the $1{,}120 \times 630$ input into an $80 \times 45$ grid of 3{,}600 tokens, and a complex embedding dimension of 192, a learned per-patch complex filter analogous to matched filtering. Because it is a shared complex convolution, it is $U(1)$-equivariant, so a global phase rotation of the input rotates every token by the same phase and the embedding gates the symmetry of the blocks above it. Such a complex patch embedding is the standard front end of complex vision transformers \citep{Eilers2023-blocks, Rempe2026-kvit, Sabokpa2026-cvt}, specialized here to pulse-echo IQ.

\paragraph{Rotary position encoding} We apply rotary position encoding (RoPE) \citep{Su2021-aj} to the queries and keys, rotating each complex token's phase by a position-dependent angle,
\begin{equation}
 z'_{n,d} = z_{n,d}\, e^{i\theta_{n,d}}, \qquad \omega_j = 10000^{-j/(D/2)},
\end{equation}
Here $d$ indexes the channel over the $D$ complex dimensions, and the rotation angle $\theta_{n,d}$ applies the token's row position to the first half of the channels and its column position to the second half, each at frequency $\omega_j$. On complex tokens, this per-channel phase rotation is identical to standard real rotate-half RoPE up to the pairing of coordinates. The same complex RoPE appears in the complex ViT of \citet{Rempe2026-kvit}. It is shared across all encoders so that positional information is matched across the architectures we compare.

\paragraph{Hermitian attention} The complex block uses Hermitian self-attention \citep{Yang2020-complextransformer, Eilers2023-blocks}. This is the natural complex score, taking the real part of the complex inner product between queries and keys:
\begin{equation}
 a_{nm} = \mathrm{softmax}_m\!\left(\frac{\mathrm{Re}\,\langle q_n, k_m\rangle}{\sqrt{d_h}}\right), \qquad \mathrm{out}_n = \sum_m a_{nm}\, v_m,
 \label{eq:hermitian-attn}
\end{equation}
where $d_h$ is the per-head complex dimension. This attention is exactly $U(1)$-equivariant, because a global phase rotation passes through the values unchanged in the attention pattern. It is implemented through a fused real kernel (Appendix~\ref{sec:appendix-flash}).

\begin{figure}[!t]
 \centering
 \includegraphics[width=\linewidth]{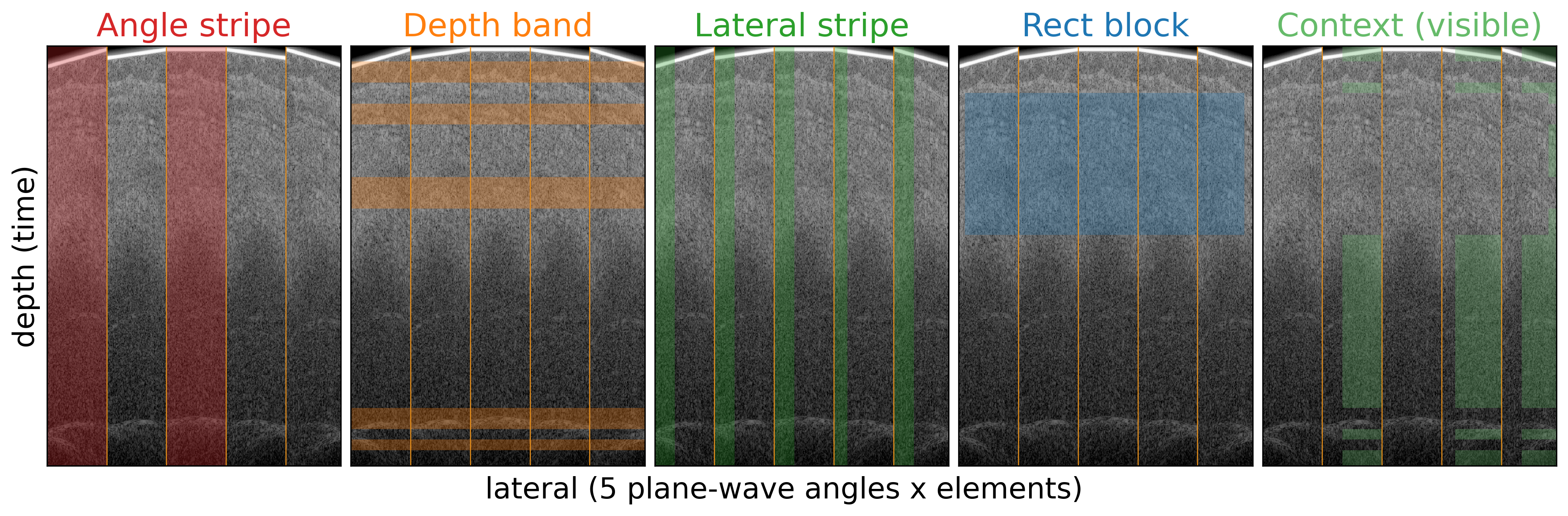}
 \caption{IQ-JEPA masking on a reshaped IQ acquisition, depth vertical and the five plane-wave angles side by side (orange lines mark the angle boundaries). The first four panels illustrate the four target-block strategies, one of which the collator draws independently for each of the four target blocks (angle stripe, depth band, lateral stripe, rectangular block). The rightmost panel is the context for this acquisition, the complement of its target union and roughly 28\% of the patches, which is the region the context encoder sees.}
 \label{fig:iqjepa-mask}
\end{figure}

\paragraph{Conjugate-product feed-forward} The feed-forward layer forms two complex projections $u$ and $w$ of each token and takes their elementwise conjugate product before normalizing and projecting out:
\begin{equation}
 p = u \odot \overline{w}, \qquad \mathrm{out} = W_{\mathrm{out}}\,\mathrm{CLN}(p).
 \label{eq:conjugate-mlp}
\end{equation}
Here $\odot$ is the elementwise product, $\overline{w}$ the complex conjugate, $\mathrm{CLN}$ complex layer normalization, and $W_{\mathrm{out}}$ the output projection. The phase of the conjugate product is the \emph{difference} of phases, $\angle p = \angle u - \angle w$. This is invariant to a global phase rotation, since $u \mapsto e^{i\phi}u$ and $w \mapsto e^{i\phi}w$ leave $u\,\overline{w}$ unchanged. This phase-difference signal is similar to the one classical coherence methods use to estimate sound speed \citep{Jaeger2015-dq}. The conjugate product is thus $U(1)$-invariant by construction.

\paragraph{Global phase augmentation} During training we rotate the whole IQ field by a random global phase,
\begin{equation}
 z \mapsto e^{i\phi} z, \qquad \phi \sim \mathcal{U}[0, 2\pi),
\end{equation}
a zero-parameter regularizer that installs the $U(1)$ invariance of the target directly in the data. This augmentation is needed because of how the data are generated. Every simulated acquisition is demodulated against the same reference, so the training set contains a single global demodulation phase and the network never sees the symmetry it is supposed to respect. Rotating the field supplies that variation. On a real system the phase varies between acquisitions, so the augmentation reproduces a source of variability that the simulation does not have. Augmentation of this kind lets a network tolerate the unseen carrier-phase rotations that otherwise break coordinate-dependent models \citep{Kumar2026-cvnnhelp}.

\subsection{Architecture}

A single complex block operates on the complex tokens. Its real and imaginary parts are then concatenated along the channel axis (doubling the width to 384) and passed to eleven standard real-valued transformer blocks (scaled dot-product attention \citep{vaswani2017attention}, GELU feed-forward, real layer normalization). These carry the ViT-S capacity through the rest of the network. The encoder has 21.9\,M parameters, about 4\% lighter than the 22.8\,M real-valued ViT it is matched against. This reduction comes from weight tying in the complex linear layers. All results below use this configuration with one complex block (total depth 12). The contribution of the single complex block, against a real-valued transformer block of the same width, is isolated in Section~\ref{sec:results:pretrain-scaling}.

\subsection{Hermitian flash attention}

Because the Hermitian score $\mathrm{Re}(Q\,K^{\mathsf{H}})$ is real-valued, complex attention reduces to a single real attention on the real and imaginary parts stacked along the feature axis. A standard fused flash-attention kernel \citep{Dao2022-flashattention} evaluates this unchanged. We adopt it because it makes the complex encoder practical to pretrain. It costs about $1.7\times$ a matched real attention layer, rather than four to eight times the cost of the explicit complex product. Appendix~\ref{sec:appendix-flash} gives the construction, the algorithm, the exactness and score-scale arguments, and the latency and memory benchmarks.

\subsection{IQ-JEPA pretraining}
\label{sec:methods:ssl}

We pretrain the encoder with a joint-embedding predictive objective adapted from I-JEPA \citep{Assran2023-ijepa}. A context encoder $f_\theta$ and a stop-gradient exponential moving average (EMA) target encoder $f_{\bar\theta}$ produce patch-level latents of the same input under different masks. A small predictor $g_\phi$, a six-layer transformer of width 384, maps the context latents to the target latents at masked positions. The loss is a smooth-$\ell_1$ distance in latent space, applied only at masked targets,
\begin{equation}
 \mathcal{L} = \sum_{j \in \text{targets}} \mathrm{smooth}\text{-}\ell_1\!\big(g_\phi(f_\theta(x_{\mathrm{ctx}}))_j,\; \mathrm{sg}[f_{\bar\theta}(x)_j]\big),
 \label{eq:jepa-loss}
\end{equation}
and the target encoder follows the context encoder by EMA,
\begin{equation}
 \bar\theta \leftarrow \tau\,\bar\theta + (1 - \tau)\,\theta.
\end{equation}
The EMA target avoids representation collapse without negative pairs. The objective matches latent representations rather than reconstructing pixels. Pixel reconstruction is poorly suited to speckle-dominated IQ, and Section~\ref{sec:results:ssl-design} compares the two objectives directly.

Masks follow the I-JEPA recipe, sampled on the reshaped IQ tensor (Figure~\ref{fig:iqjepa-mask}). Each sample draws four target blocks, each independently choosing one of four axis-aligned strategies (angle stripes, depth bands, lateral stripes, and rectangular blocks) with equal probability. The context is the complement of their union, roughly 28\% of the patches. Because the reshape places depth on one axis and the (angle, element) structure on the other, target blocks span depths, lateral elements, or occasionally transmit angles by aspect ratio. The objective therefore averages cross-depth, cross-element, and cross-angle prediction. Context and rotary indices are gathered per sample. The context mask is the complement of one sample's target union and is shared across the batch, following the reference I-JEPA collator, so for the other samples in a batch about 20\% of their own target patches stay visible to the context encoder. Pretraining draws from the 63{,}435 training acquisitions with no labels. Of these, 90\% (57{,}091) supply the gradient steps and 10\% (6{,}344) are held out as a self-supervised validation set used only for checkpoint selection. We carve this validation set out of the training split rather than reuse the fine-tuning validation set. A pretraining checkpoint is therefore never selected on data the model is later scored against, which keeps the encoder blind to the validation and test acquisitions at every stage. The optimizer is AdamW, and the learning rate warms up and then decays on a cosine schedule. Pretraining lasts 100 epochs and runs under distributed data parallelism on eight H100 GPUs (two nodes of four), at a per-GPU batch of 16 and an effective batch of 128. Training completes in about six hours. The full training and architecture hyperparameters are in Appendix~\ref{sec:appendix-impl} (Table~\ref{tab:hyperparameters}).

\subsection{Downstream fine-tuning}
\label{sec:methods:finetune}

For the downstream task, we discard $g_\phi$ and $f_{\bar\theta}$. We attach a lightweight convolutional regression head to the encoder's final tokens, reshaped to the patch grid. It is a batch normalization, a $3\times3$ convolution to 64 channels, a GELU, and a $1\times1$ convolution to the single output channel, with the output bilinearly upsampled to the sound-speed grid. Appendix~\ref{sec:appendix-finetune-projection} (Figure~\ref{fig:finetune-projection}) shows how fine-tuning fuses the channel data into the spatial domain. We fine-tune encoder and head jointly with a normalized mean-squared-error loss against the Fullwave sound-speed ground truth, where the target is z-scored with fixed constants and predictions are inverted with the same constants before scoring. The optimizer is AdamW, the learning rate warms up and then decays on a cosine schedule, and we train for 100 epochs. Two scaling protocols are used. The \emph{label-count} study fixes the pretraining set and varies the labeled fine-tune fraction over $\{1{,}000;\, 5{,}000;\, 10{,}000;\, 63{,}435\}$. These labeled subsets are nested prefixes of a single fixed shuffle of the training split, so they are identical across the supervised and self-supervised conditions. The \emph{pretraining-size} study fixes the labeled fraction at 10{,}000 and varies the unlabeled set over $\{0,\, 2{,}000,\, 5{,}000,\, 10{,}000,\, 20{,}000,\, 40{,}000,\, 63{,}435\}$, where $0$ is supervised from scratch. A second-property study fine-tunes the same self-supervised pretrained encoder on attenuation labels, with the pretraining shared and not repeated. This study is reported in Section~\ref{sec:results:attenuation}.

\subsection{Baselines and evaluation}
\label{sec:methods:baselines}

We compare the estimators at matched model scale. Global phase augmentation acts on the IQ input, so it is applied identically to the two IQ transformer estimators, the real-valued ViT and the Hermitian ViT, and their comparison is augmentation-matched. It does not apply to the RF-input estimators, where the global demodulation phase is not a free parameter. The transformer estimators share the identical rotary position encoding. The comparison therefore isolates the input representation, holding augmentation and positional encoding fixed. The real-valued ViT is a different backbone, so the matched comparison of a complex block against a real block is the single-block ablation (Figure~\ref{fig:noblock}). InversionNet \citep{Wu2019-inversionnet} is a convolutional inverse-mapping baseline, run on RF. Its architecture is detailed in Appendix~\ref{sec:appendix-inversionnet}. A real-valued ViT (the OpenFWI Kaggle architecture of Section~\ref{sec:methods:encoder}) is run supervised on RF and on IQ. This isolates the effect of the input modality, with the encoder held real. The real-valued ViT takes the IQ input as its in-phase and quadrature parts stacked as two real channels, whereas the Hermitian ViT operates on the complex signal directly. The Hermitian ViT is run both supervised (no pretraining) and with IQ-JEPA pretraining followed by fine-tuning. This isolates the effect of self-supervision, with the encoder held fixed. The same I-JEPA objective is also applied to the real-valued ViT, giving an SSL versus SSL architecture comparison.

The primary metric is the mean absolute error (MAE) of predicted sound speed in meters per second, on a held-out test set. It is the pixel-wise mean absolute error over the full $512 \times 256$ sound-speed grid, averaged over all test acquisitions, with no region excluded, and the prediction is bilinearly upsampled to this grid before scoring. Every row of Table~\ref{tab:label-efficiency} is a three-seed mean over seeds 42, 123, and 7. Across the three headline seeds the test set is held fixed while the training and validation split is reshuffled, so the reported standard deviation includes label-sampling variance. The standard deviation across the three seeds is at most 1.6\ms{}, at the 1{,}000-label real-valued ViT RF point, and under 1.3\ms{} at every other cell, far below the self-supervised error reductions of 2.80 to 14.97\ms{} across the label range. Seed choice therefore does not change the conclusions. For pretraining, we keep the checkpoint with the lowest self-supervised validation loss, and for fine-tuning the one with the lowest validation MAE. The selected checkpoints for the full-corpus pretraining runs all fall after epoch 90 of 100, so this selection does not latch an early collapse. The frozen self-supervised features show rich structure rather than collapse (Figure~\ref{fig:learned-features}). All comparisons are at matched label fractions and identical optimizer hyperparameters, with 100-epoch schedules unless stated otherwise. %
\section{Results}
\label{sec:results}

\subsection{Label efficiency under self-supervision}
\label{sec:results:label-efficiency}

\begin{figure}[tb]
 \centering
 \begin{subfigure}[t]{0.49\textwidth}
 \centering
 \begingroup
 \renewcommand{\tiny}{\fontsize{6.5}{7.8}\selectfont}%
 \renewcommand{\scriptsize}{\fontsize{9.1}{10.9}\selectfont}%
 \renewcommand{\footnotesize}{\fontsize{10.4}{12.5}\selectfont}%
 \renewcommand{\small}{\fontsize{11.7}{14}\selectfont}%
 \renewcommand{\normalsize}{\fontsize{13}{15.6}\selectfont}%
 \adjustbox{max width=\linewidth}{%
\begin{tikzpicture}
  \begin{axis}[
      iqaxis, name=ax,
      width=9cm, height=7cm,
      xlabel={Label count (log)},
      ylabel={Test MAE (m/s)},
      ymin=0, ymax=55,
      ytick={0,10,20,30,40,50},
      xmode=log, log basis x=10,
      log ticks with fixed point,
      xminorticks=false,
      xtick={1000,5000,10000,63435},
      xticklabels={1K,5K,10K,63K},
      enlarge x limits=0.06,
      legend style={font=\scriptsize, draw=black!25, fill=white,
          at={(0.5,-0.27)}, anchor=north, legend columns=1,
          cells={anchor=west}},
    ]
    \addplot[iq/cnn, error bars/.cd, y dir=both, y explicit,
        error bar style={line width=0.9pt, solid}, error mark options={mark size=3.5pt, line width=0.9pt, solid}]
      coordinates {(1000,51.38) +- (0,0.11) (5000,39.49) +- (0,0.74) (10000,34.73) +- (0,0.07) (63435,18.32) +- (0,0.76)};
    \addlegendentry{InversionNet (RF), supervised}
    \addplot[iq/c1sup, error bars/.cd, y dir=both, y explicit,
        error bar style={line width=0.9pt, solid}, error mark options={mark size=3.5pt, line width=0.9pt, solid}]
      coordinates {(1000,47.26) +- (0,0.43) (5000,36.32) +- (0,1.21) (10000,30.04) +- (0,0.34) (63435,15.01) +- (0,0.13)};
    \addlegendentry{Real-valued ViT (IQ), supervised}
    \addplot[iq/c5sup, error bars/.cd, y dir=both, y explicit,
        error bar style={line width=0.9pt, solid}, error mark options={mark size=3.5pt, line width=0.9pt, solid}]
      coordinates {(1000,45.44) +- (0,0.72) (5000,29.87) +- (0,1.24) (10000,23.46) +- (0,0.18) (63435,11.51) +- (0,0.15)};
    \addlegendentry{Hermitian ViT (IQ), supervised}
    \addplot[iq/realssl, error bars/.cd, y dir=both, y explicit,
        error bar style={line width=0.9pt, solid}, error mark options={mark size=3.5pt, line width=0.9pt, solid}]
      coordinates {(1000,33.93) +- (0,0.93) (5000,22.76) +- (0,0.27) (10000,18.21) +- (0,0.08) (63435,9.99) +- (0,0.09)};
    \addlegendentry{Real-valued ViT (IQ), IQ-JEPA SSL+ft}
    \addplot[iq/winner, error bars/.cd, y dir=both, y explicit,
        error bar style={line width=0.9pt}, error mark options={mark size=3.5pt, line width=0.9pt}]
      coordinates {(1000,30.47) +- (0,0.91) (5000,19.72) +- (0,0.35) (10000,15.60) +- (0,0.26) (63435,8.71) +- (0,0.17)};
    \addlegendentry{Hermitian ViT (IQ), IQ-JEPA SSL+ft (proposed)}

    \node[iqred, font=\scriptsize, anchor=north east, yshift=-1pt] at (axis cs:10000,15.60) {15.60};
    \node[iqred, font=\scriptsize, anchor=north east, yshift=-1pt] at (axis cs:63435,8.71) {8.71};
    \coordinate (BBSW) at (rel axis cs:0,0);
    \coordinate (BBNE) at (rel axis cs:1,1);
  \end{axis}
  \useasboundingbox ([shift={(-1.5cm,-4.7cm)}]BBSW) rectangle ([shift={(0.4cm,0.35cm)}]BBNE);
\end{tikzpicture}}%
 \endgroup

 \caption{Label efficiency. Sound-speed test MAE versus label count for five of the six estimators in Table~\ref{tab:label-efficiency} (the RF-supervised real-valued ViT is omitted, as it nearly coincides with its IQ counterpart), under supervised training and under IQ-JEPA pretraining then fine-tuning. The IQ-JEPA Hermitian ViT reaches 15.60\ms{} at 10{,}000 labels against 23.46\ms{} supervised, a roughly threefold label reduction, and is lowest at every count. Error bars are the three-seed standard deviation on every curve. For the IQ-JEPA Hermitian ViT the spread shrinks from 0.91\ms{} at 1{,}000 labels to 0.17\ms{} at the full count, and is smaller than the marker at the higher counts.}
 \label{fig:label-efficiency}
 \end{subfigure}
 \hfill
 \begin{subfigure}[t]{0.49\textwidth}
 \centering
 \begingroup
 \renewcommand{\tiny}{\fontsize{6.5}{7.8}\selectfont}%
 \renewcommand{\scriptsize}{\fontsize{9.1}{10.9}\selectfont}%
 \renewcommand{\footnotesize}{\fontsize{10.4}{12.5}\selectfont}%
 \renewcommand{\small}{\fontsize{11.7}{14}\selectfont}%
 \renewcommand{\normalsize}{\fontsize{13}{15.6}\selectfont}%
 \adjustbox{max width=\linewidth}{%
\begin{tikzpicture}
  \begin{axis}[
      iqaxis,
      width=9cm, height=7cm,
      xmode=log, log basis x=10,
      ymode=log, log basis y=10,
      log ticks with fixed point,
      xminorticks=false, yminorticks=false,
      xtick={2000,5000,10000,20000,40000,63435},
      xticklabels={2K,5K,10K,20K,40K,63K},
      xmin=1650, xmax=82000,
      xlabel={Unlabeled pretraining set (log)},
      ytick={15,20,25,30},
      yticklabels={15,20,25,30},
      ymin=14, ymax=31.5,
      ylabel={Test MAE at 10K labels, m/s (log)},
      legend style={font=\scriptsize, draw=black!25, fill=white,
          at={(0.5,-0.27)}, anchor=north, legend columns=1,
          cells={anchor=west}},
      reverse legend,
    ]
    \draw[iqgray, densely dotted] (axis cs:10000,14) -- (axis cs:10000,31.5);
    \node[iqgray, font=\scriptsize, anchor=south west, rotate=90]
    at (axis cs:10000,14.3) {10K fine-tuning labels};

    \addplot[iqred, densely dashed, line width=0.7pt, no marks, forget plot]
    coordinates {(10000,20.85) (63435,15.43)};
    \addplot[iqgreen, densely dashed, line width=0.7pt, no marks, forget plot]
    coordinates {(10000,27.46) (63435,18.48)};

    \addplot[iq/winner, error bars/.cd, y dir=both, y explicit,
      error bar style={line width=0.8pt}, error mark options={mark size=2.5pt, line width=0.8pt}]
    coordinates {
        (2000,22.20) +- (0,0.68) (5000,21.47) +- (0,0.29) (10000,20.81) +- (0,0.15)
        (20000,18.82) +- (0,0.35) (40000,16.31) +- (0,0.27) (63435,15.60) +- (0,0.26)
      };
    \addlegendentry{Hermitian ViT (IQ)}

    \addplot[iq/realssl, error bars/.cd, y dir=both, y explicit,
      error bar style={line width=0.8pt}, error mark options={mark size=2.5pt, line width=0.8pt}]
    coordinates {
        (2000,28.78) +- (0,0.57) (5000,28.14) +- (0,0.45) (10000,27.17) +- (0,0.95)
        (20000,23.93) +- (0,0.69) (40000,20.70) +- (0,0.46) (63435,18.21) +- (0,0.08)
      };
    \addlegendentry{Real-valued ViT (IQ)}

    \node[anchor=west, font=\scriptsize, text=iqgreen] at (axis cs:23000,25.4) {$b\approx0.21$};
    \node[anchor=north east, font=\scriptsize, text=iqred] at (axis cs:40000,16.4) {$b\approx0.16$};
    \node[anchor=south west, font=\scriptsize, text=iqred] at (axis cs:63435,15.60) {15.60};
    \node[anchor=south west, font=\scriptsize, text=iqgreen] at (axis cs:63435,18.21) {18.21};
    \coordinate (BBSW) at (rel axis cs:0,0);
    \coordinate (BBNE) at (rel axis cs:1,1);
  \end{axis}
  \useasboundingbox ([shift={(-1.5cm,-4.7cm)}]BBSW) rectangle ([shift={(0.4cm,0.35cm)}]BBNE);
\end{tikzpicture}}%
 \endgroup

 \caption{Pretraining-data scaling for two IQ encoders, with the fine-tuning set fixed at 10{,}000 labels (dotted line), on log-log axes. Error is flat below that count, then decreases steadily with the pretraining size above it, with descriptive scaling exponents (dashed) of 0.16 for the Hermitian ViT and a steeper 0.21 for the real-valued ViT. The Hermitian ViT is lowest at every pretraining size, from 22.20 to 15.60\ms{}, but the gap narrows as the pretraining set grows. Error bars are the three-seed standard deviation on both curves.}
 \label{fig:pretrain-scaling}
 \end{subfigure}
 \caption{Scaling behavior of self-supervised sound-speed estimation on the held-out Fullwave plane-wave test split. Panel (a) shows label efficiency at fixed pretraining, and panel (b) shows pretraining-data scaling at a fixed label count.}
 \label{fig:scaling}
\end{figure}
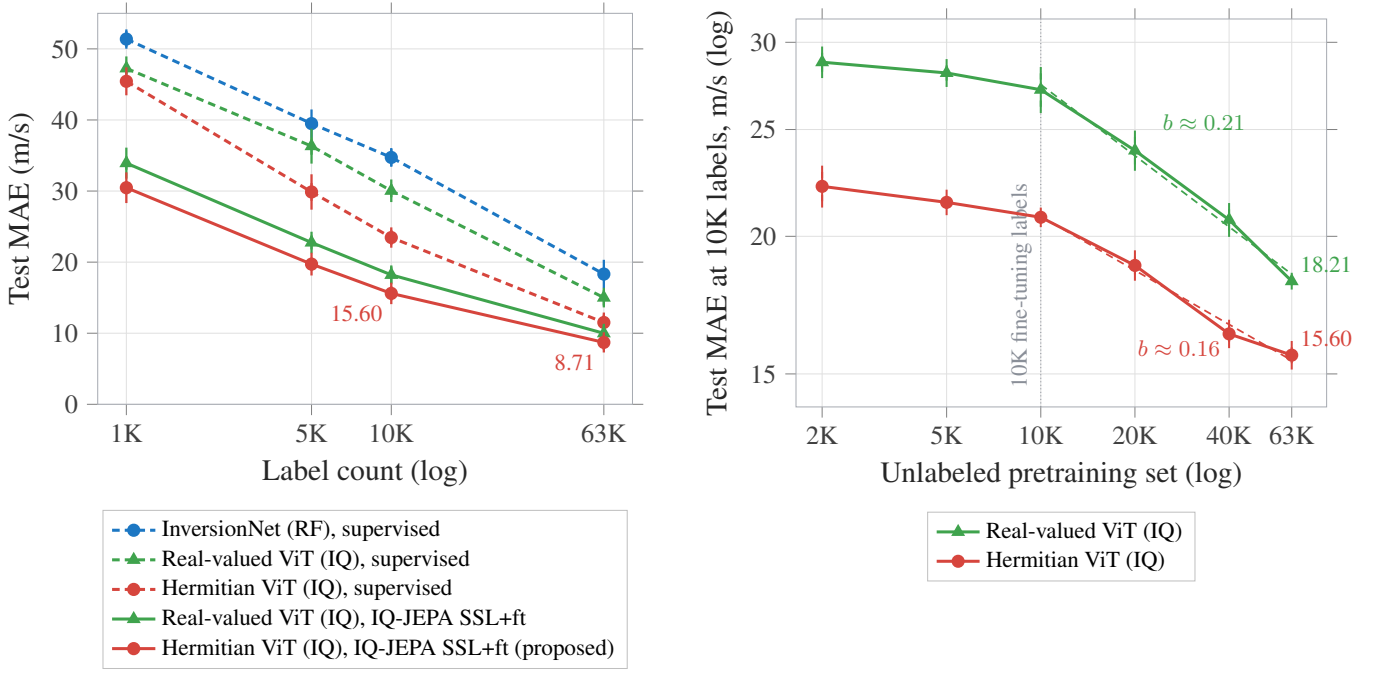

Self-supervision improves label efficiency. Table~\ref{tab:label-efficiency} and Figure~\ref{fig:label-efficiency} report sound-speed test MAE across four label counts for the estimators of Section~\ref{sec:methods:baselines}. Self-supervised pretraining followed by fine-tuning (the IQ-JEPA family) lies below supervised training at every label count, and the Hermitian ViT is lowest throughout.

\begin{widetable}[tb]
 \caption{Sound-speed test MAE (m/s) on held-out Fullwave plane-wave data versus fine-tuning label count, lower is better. Every row is a three-seed mean $\pm$ standard deviation over seeds 42, 123, and 7, and the proposed model is in bold. IQ-JEPA is self-supervised pretraining on 63{,}435 unlabeled acquisitions then fine-tuning. Mean-predictor floor 74.5\ms{}. }
 \label{tab:label-efficiency}
 \centering
 \small
 \begin{tabular}{lllcccc}
 \toprule
 Encoder & Input & Training & 1K & 5K & 10K & 63K \\
 \midrule
 InversionNet & RF & supervised & $51.38\pm0.11$ & $39.49\pm0.74$ & $34.73\pm0.07$ & $18.32\pm0.76$ \\
 Real-valued ViT & RF & supervised & $46.89\pm1.56$ & $35.87\pm0.45$ & $30.65\pm0.28$ & $13.95\pm0.27$ \\
 Real-valued ViT & IQ & supervised & $47.26\pm0.43$ & $36.32\pm1.21$ & $30.04\pm0.34$ & $15.01\pm0.13$ \\
 Hermitian ViT & IQ & supervised & $45.44\pm0.72$ & $29.87\pm1.24$ & $23.46\pm0.18$ & $11.51\pm0.15$ \\
 Real-valued ViT & IQ & IQ-JEPA SSL+ft & $33.93\pm0.93$ & $22.76\pm0.27$ & $18.21\pm0.08$ & $9.99\pm0.09$ \\
 \textbf{Hermitian ViT} & \textbf{IQ} & \textbf{IQ-JEPA SSL+ft} & $\bm{30.47\pm0.91}$ & $\bm{19.72\pm0.35}$ & $\bm{15.60\pm0.26}$ & $\bm{8.71\pm0.17}$ \\
 \bottomrule
 \end{tabular}
\end{widetable}

Against the matched supervised model, IQ-JEPA at 10{,}000 labels reaches 15.60\ms{} (three-seed mean, Table~\ref{tab:label-efficiency}). The same Hermitian ViT trained from scratch reaches 23.46\ms, a 7.86\ms{} reduction. Supervised training does not reach 15.60\ms{} until roughly three times as many labels, about a threefold gain in label efficiency at fixed accuracy. The saving is largest where labels are scarcest, more than fourfold at 1{,}000 labels. This gain is reached at only about a one-to-six ratio of labeled to unlabeled data (10{,}000 labels against the 63{,}435-acquisition set). This ratio is far below those typical of self-supervised pretraining. Because the pretraining-data scaling does not saturate (Figure~\ref{fig:pretrain-scaling}), label efficiency should grow further as the unlabeled data expands. Against the convolutional baseline, IQ-JEPA is about $2.2\times$ lower than InversionNet at 10{,}000 labels (15.60 versus 34.73\ms). The supervised Hermitian ViT alone reaches 23.46\ms{} (about $1.5\times$ over InversionNet, from the architecture and the complex encoder), and self-supervision takes it the rest of the way to 15.60\ms{}. The complex encoder keeps an edge even in the self-supervised regime at the largest label count. With the same IQ-JEPA pretraining and fine-tuning, the Hermitian ViT reaches $8.71\pm0.17\ms{}$ at 63{,}435 labels against $9.99\pm0.09\ms{}$ for the real-valued ViT. This 1.28\ms{} margin is far outside the seed spread and persists where the supervised gap has nearly closed. Supervised IQ training with a real-valued ViT reaches about 30\ms{} at 10{,}000 labels, close to the RF supervised ViT. The IQ input alone therefore does not improve supervised accuracy. The complex encoder and the self-supervised objective are what lower the error. Sound speed carries clinical value in several settings. It separates malignant from benign breast lesions by about 35\ms{} \citep{Li2009-breastust}. It tracks liver fat at roughly 1.7\ms{} per percent and about 10\ms{} per steatosis grade \citep{Telichko2022-gz}. It also sets pulse-echo focusing quality, since a wrong value aberrates the beam \citep{Ali2022-wq}. Against these scales, the per-pixel error here is well below the breast-lesion separation and on the order of one liver steatosis grade, 8.71\ms{} at the full label count and 15.60\ms{} at 10{,}000 labels. Pulse-echo physics estimators are accurate in simple layered media where lateral variation is negligible \citep{Ali2022-wq, Stahli2020-fn}, but their absolute accuracy on the two-dimensional heterogeneity of tissue in vivo is rarely reported. The errors here are absolute MAE on heterogeneous layered and inclusion phantoms.

\begin{figure}[tb]
 \centering
 \includegraphics[width=0.8\linewidth]{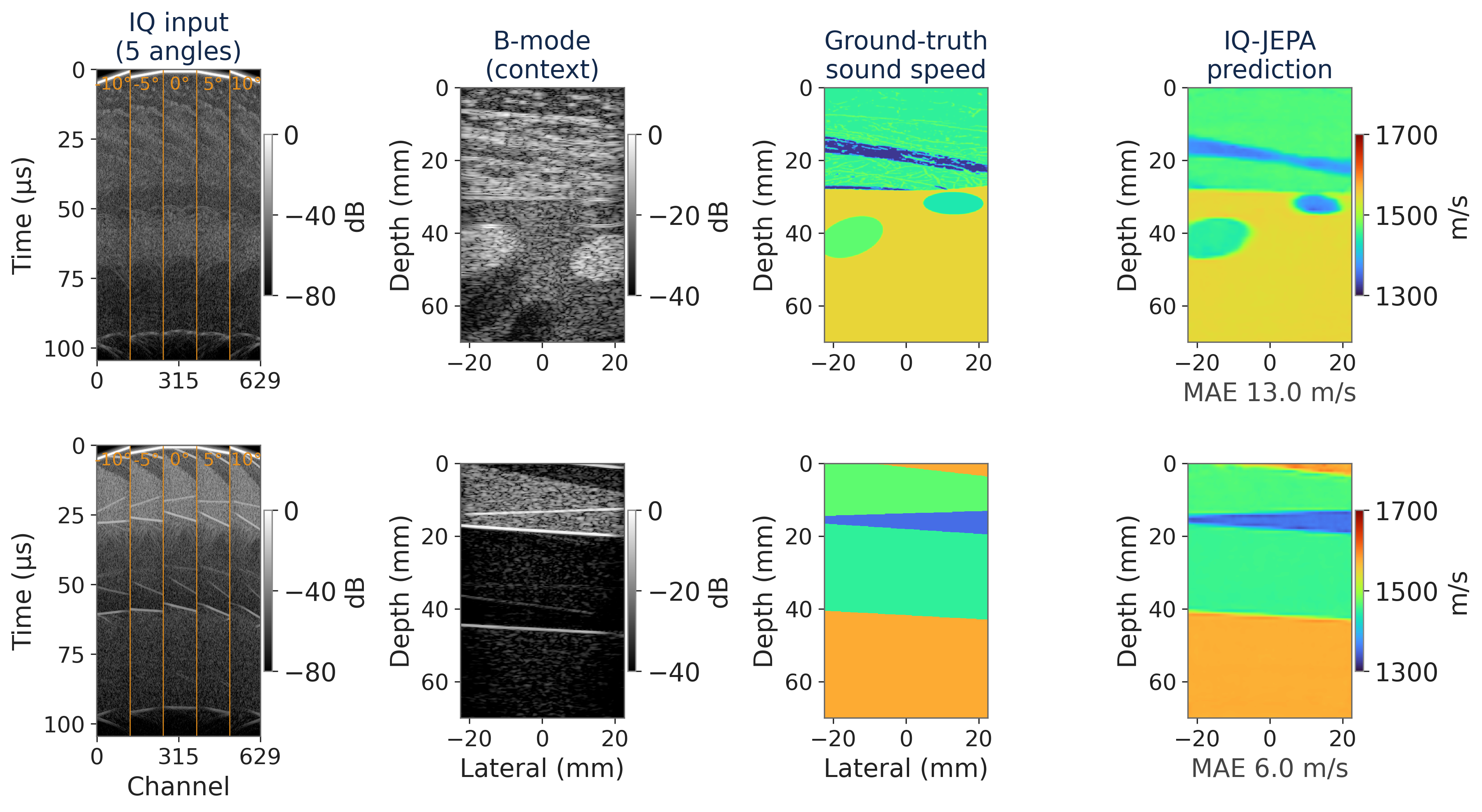}
 \caption{Qualitative sound-speed predictions from the IQ-JEPA Hermitian ViT on two held-out samples, an abdominal phantom with an inclusion (top) and a random 5-layer phantom (bottom). Each row shows the IQ input, the B-mode, the ground-truth sound speed, and the prediction, which recovers the layer boundaries, inclusions, and speed ordering. Further predictions are in Appendix~\ref{sec:appendix-gallery}.}
 \label{fig:sos-predictions}
\end{figure}

\begin{figure}[tb]
 \centering
 \includegraphics[width=0.8\linewidth]{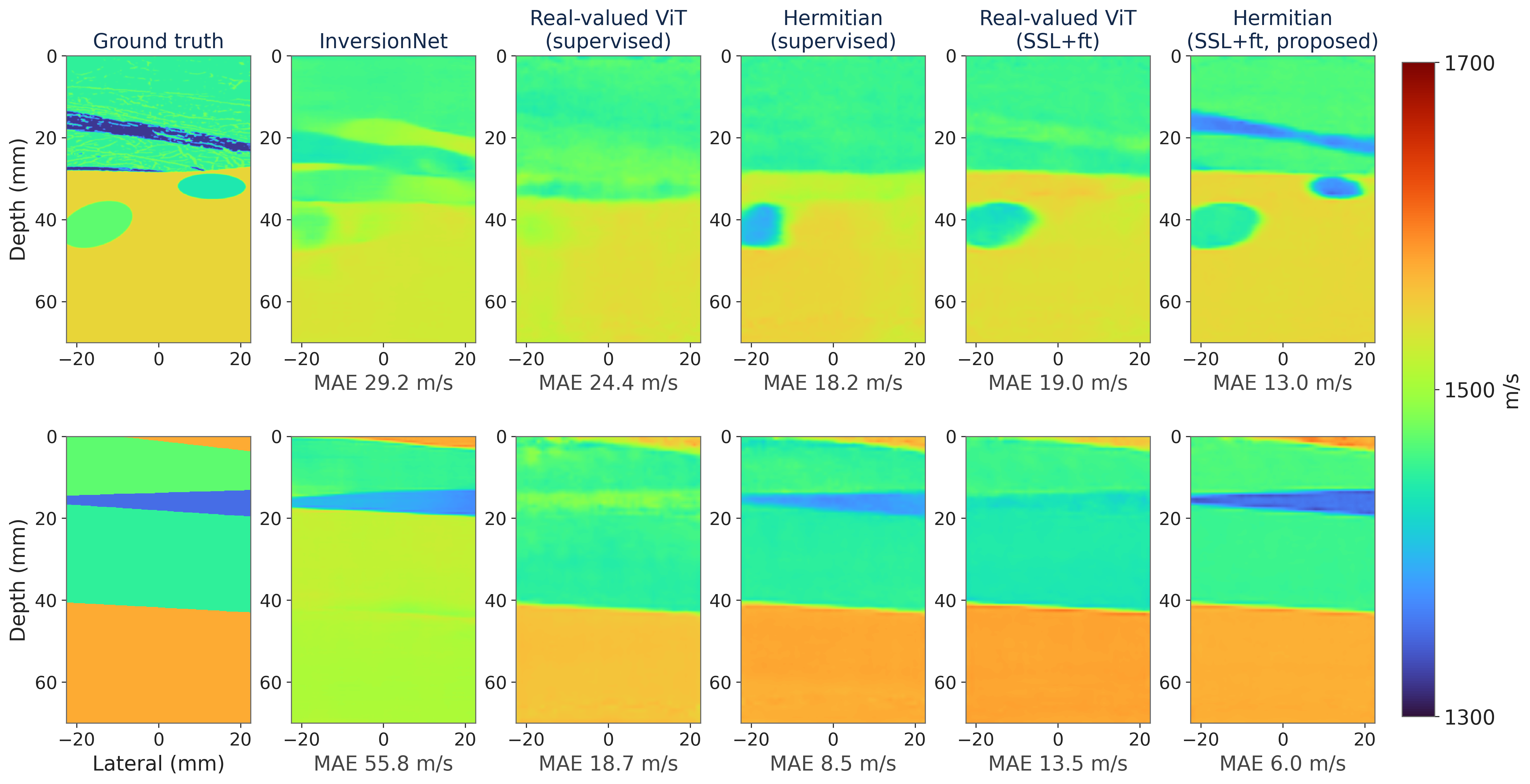}
 \caption{Inter-model comparison at 63{,}435 labels on two held-out samples, an abdominal phantom with an inclusion (top) and a random 5-layer phantom (bottom). Columns are ground truth, InversionNet, the supervised real and Hermitian ViTs, the self-supervised real-valued ViT, and the IQ-JEPA Hermitian ViT, with per-sample MAE. The complex models hold thin layers the convolutional baseline smooths over, and only the IQ-JEPA Hermitian ViT recovers the inclusion. More comparisons are in Appendix~\ref{sec:appendix-intermodel}.}
 \label{fig:model-comparison}
\end{figure}

Qualitative predictions on held-out samples recover the layer boundaries, inclusions, and speed ordering of the ground truth across phantom families (Figure~\ref{fig:sos-predictions}). At the largest label count, a model comparison shows the complex Hermitian models holding thin layers where the convolutional baseline collapses them (Figure~\ref{fig:model-comparison}). Only the self-supervised IQ-JEPA Hermitian ViT recovers the inclusion. The other three models, InversionNet and the supervised real and Hermitian ViTs, smooth it away.

\subsection{A pretraining-data scaling trend}
\label{sec:results:pretrain-scaling}

We fix the labeled fine-tune set at 10{,}000 and grow only the unlabeled pretraining set. Sound-speed error for the Hermitian ViT then falls from 23.46\ms{} with no pretraining to 22.20, 21.47, 20.81, 18.82, 16.31, and 15.60\ms{} at pretraining sizes of 2{,}000, 5{,}000, 10{,}000, 20{,}000, 40{,}000, and 63{,}435 (Figure~\ref{fig:pretrain-scaling}). The curve does not follow a single scaling exponent. It is nearly flat on the log-log axes below the 10{,}000-sample fine-tuning set, where a fivefold larger pretraining set from 2{,}000 to 10{,}000 yields only 1.39\ms{}. It steepens once the pretraining set grows past that size, and above 10{,}000 the error decreases steadily. A straight-line fit on the log-log axes gives a descriptive exponent of 0.16. The elbow falls near the size of the 10{,}000-label fine-tuning set. One reading is that while the pretraining set stays smaller than the labeled set the fine-tuning data dominates, and that unlabeled data begins to pay off once it carries more examples than the labeled set supplies. The error is still descending at the largest pretraining set we generated, though the rate is slowing.

The real-valued ViT on the same IQ shows the same regime structure, flat below 10{,}000 samples and steep above it. It stays above the Hermitian ViT at every pretraining size (30.04\ms{} supervised, then 28.78, 28.14, 27.17, 23.93, 20.70, and 18.21\ms{} across this range). Fitting the productive regime gives an exponent of 0.16 for the Hermitian ViT and a steeper 0.21 for the real-valued ViT. So the real encoder closes the gap as the data grows. The Hermitian margin falls from 6.4\ms{} at 10{,}000 to 2.6\ms{} at 63{,}435. The complex block therefore has a data-efficiency advantage. It is largest when unlabeled data is scarce and shrinks with scale, consistent with the single-block ablation of Figure~\ref{fig:noblock}. Both curves are three-seed means at every pretraining size and carry seed error bars, with a standard deviation of 0.15 to 0.68\ms{} for the Hermitian ViT and 0.08 to 0.95\ms{} for the real-valued ViT. In the intended application the in vivo or ex vivo IQ is abundant and unlabeled, while the labeled sound-speed maps come from a small in silico set.

Figure~\ref{fig:noblock} isolates the single complex block. Its value grows with pretraining and is largest at a small unlabeled set. Replacing the block (Hermitian attention and conjugate-product feed-forward together) with a real-valued block raises error by 3.12\ms{} at a 10{,}000-sample pretraining set (95\% CI 2.39 to 3.84) and by 1.95\ms{} at 40{,}000 (0.63 to 3.27). This holds at matched width and a slightly larger parameter count (22.77\,M versus 21.88\,M). Without the block the encoder does not improve from supervised to a 10{,}000-sample pretraining set (23.27 to 23.93\ms{}), while the block lets it gain 2.65\ms{} over the same range. The block is what turns a small unlabeled set into accuracy. At the two ends of the range the measurement does not resolve a difference. With no pretraining the two configurations differ by $-0.19\ms{}$ (CI $-2.64$ to $2.26$) and at the full pretraining set by $0.56\ms{}$ (CI $-0.02$ to $1.14$). The complex block is a data-efficiency factor, secondary to the pretraining size and the self-supervised objective.

\begin{narrowfigure}[tb]
 \centering
 \begingroup
 \renewcommand{\tiny}{\fontsize{6.5}{7.8}\selectfont}%
 \renewcommand{\scriptsize}{\fontsize{9.1}{10.9}\selectfont}%
 \renewcommand{\footnotesize}{\fontsize{10.4}{12.5}\selectfont}%
 \renewcommand{\small}{\fontsize{11.7}{14}\selectfont}%
 \renewcommand{\normalsize}{\fontsize{13}{15.6}\selectfont}%
 \adjustbox{max width=\linewidth}{%
\begin{tikzpicture}
  \begin{axis}[
      iqaxis,
      width=7cm, height=5.4cm,
      xmode=log, log basis x=10,
      xminorticks=false,
      xtick={10000,40000,63435},
      xticklabels={10K,40K,63K},
      xmin=8000, xmax=90000,
      xlabel={unlabeled pretraining set (log)},
      ylabel={Test MAE at 10K labels (m/s)},
      ymin=13, ymax=26,
      legend style={font=\footnotesize, draw=black!25, fill=white,
          at={(0.5,-0.42)}, anchor=north, legend columns=1, cells={anchor=west}},
    ]
    \addplot[iqgreen, densely dotted, line width=0.8pt, no marks, forget plot] coordinates {(8000,23.27) (90000,23.27)};
    \addplot[iqred, densely dotted, line width=0.8pt, no marks, forget plot] coordinates {(8000,23.46) (90000,23.46)};
    \node[anchor=north west, font=\scriptsize, text=iqgray] at (axis cs:8000,25.8)
    {dotted: supervised (no pretraining)};

    \addplot[iqgreen, mark=square*, densely dashed, error bars/.cd, y dir=both, y explicit,
      error bar style={line width=0.8pt, solid}, error mark options={mark size=2.5pt, line width=0.8pt, solid}]
    coordinates {
        (10000,23.93) +- (0,0.34) (40000,18.26) +- (0,0.62) (63435,16.16) +- (0,0.07)
      };
    \addlegendentry{Real-valued block (matched width)}

    \addplot[iq/winner, error bars/.cd, y dir=both, y explicit,
      error bar style={line width=0.8pt}, error mark options={mark size=2.5pt, line width=0.8pt}]
    coordinates {
        (10000,20.81) +- (0,0.15) (40000,16.31) +- (0,0.27) (63435,15.60) +- (0,0.26)
      };
    \addlegendentry{Hermitian ViT (with complex block)}

    \node[anchor=north west, font=\scriptsize, text=iqred]  at (axis cs:63435,15.60) {15.60};
  \end{axis}
\end{tikzpicture}}%
 \endgroup

 \caption{One complex-valued block versus a real-valued block, at matched complex patch embedding and width, across pretraining size. The block helps most at a small pretraining set, and the two configurations are not separated at either end of the range. Points are seed means with error bars, three-seed on both curves at every pretraining size, and the dotted no-pretraining baselines are three-seed means.}
 \label{fig:noblock}
\end{narrowfigure}

\subsection{A transferable representation}
\label{sec:results:transfer}

The self-supervised representation transfers in three settings. Its frozen features already expose sound speed to a simple probe. The same encoder fine-tunes to attenuation, a distinct property. Finally, the representation carries across phantom distributions without new pretraining.

\subsubsection{Probing the frozen representation}
The self-supervised objective uses no labels, yet its frozen features already encode sound speed. We freeze the encoder and train only a probe head at 10{,}000 labels, sweeping its capacity from a linear map to an attention head. The control is the same probes on a randomly initialized encoder, which isolates decoder capacity (Appendix~\ref{sec:appendix-probe}, Figure~\ref{fig:appendix-probe-ladder}). At each head capacity, the self-supervised features read out better than random. For example, they reach 45.64 versus 70.29\ms{} at a convolutional head. The accessible sound-speed structure is therefore a property of the representation, not of the probe head. The margin traces an inverted-U over head capacity and peaks at intermediate capacity, where the readout depends on the representation rather than on the head. It narrows at both ends. This is only because a full-attention head performs its own multi-angle fusion and so leans less on any pretrained encoder, not because the representation is uninformative there.

\subsubsection{Transfer to attenuation}
\label{sec:results:attenuation}
\begin{figure}[tb]
 \centering
 \begin{subfigure}[b]{0.35\textwidth}
 \centering
 \begingroup
 \renewcommand{\tiny}{\fontsize{6.5}{7.8}\selectfont}%
 \renewcommand{\scriptsize}{\fontsize{9.1}{10.9}\selectfont}%
 \renewcommand{\footnotesize}{\fontsize{10.4}{12.5}\selectfont}%
 \renewcommand{\small}{\fontsize{11.7}{14}\selectfont}%
 \renewcommand{\normalsize}{\fontsize{13}{15.6}\selectfont}%
 \adjustbox{max width=\linewidth}{%
\begin{tikzpicture}
  \begin{axis}[
      iqaxis,
      width=9cm, height=7cm,
      xmode=log, log basis x=10,
      log ticks with fixed point,
      xminorticks=false,
      xtick={1000,5000,10000,63435},
      xticklabels={1K,5K,10K,63K},
      enlarge x limits=0.06,
      xlabel={Label count (log)},
      ylabel={Attenuation test MAE (dB/cm/MHz)},
      ymin=0.04, ymax=0.132,
      ytick={0.04,0.06,0.08,0.10,0.12},
      yticklabels={0.04,0.06,0.08,0.10,0.12},
      legend style={font=\scriptsize, draw=black!25, fill=white,
          at={(0.5,-0.27)}, anchor=north, legend columns=1, cells={anchor=west}},
    ]
    \addplot[gray!55, densely dashed, no marks, forget plot] coordinates {(1000,0.118)(63435,0.118)};
    \node[gray!75, font=\scriptsize, anchor=north east] at (axis cs:63435,0.118) {mean predictor 0.118};
    \addplot[gray!55, densely dotted, no marks, forget plot] coordinates {(1000,0.093)(63435,0.093)};
    \node[gray!75, font=\scriptsize, anchor=south east] at (axis cs:63435,0.093) {frozen probe 0.093};

    \addplot[iqgreen, mark=square*, densely dashed] coordinates {(1000,0.0916)(5000,0.0789)(10000,0.0699)(63435,0.0536)};
    \addlegendentry{Supervised (from scratch)}
    \addplot[iq/winner] coordinates {(1000,0.0776)(5000,0.0598)(10000,0.0552)(63435,0.0461)};
    \addlegendentry{IQ-JEPA SSL+ft (proposed)}

    \node[iqred, font=\scriptsize, anchor=north east] at (axis cs:10000,0.0552) {0.0552};
    \node[iqred, font=\scriptsize, anchor=north east] at (axis cs:63435,0.0461) {0.0461};
    \node[iqgreen, font=\scriptsize, anchor=south, yshift=1pt] at (axis cs:63435,0.0536) {0.0536};
  \end{axis}
\end{tikzpicture}}%
 \endgroup

 \caption{Test MAE versus label count. Self-supervised pretraining then fine-tuning is below supervised training from scratch at every label count. The dashed line is the mean-predictor floor, and the dotted line is the frozen self-supervised probe.}
 \label{fig:attenuation}
 \end{subfigure}
 \hfill
 \begin{subfigure}[b]{0.63\textwidth}
 \centering
 \includegraphics[width=\linewidth]{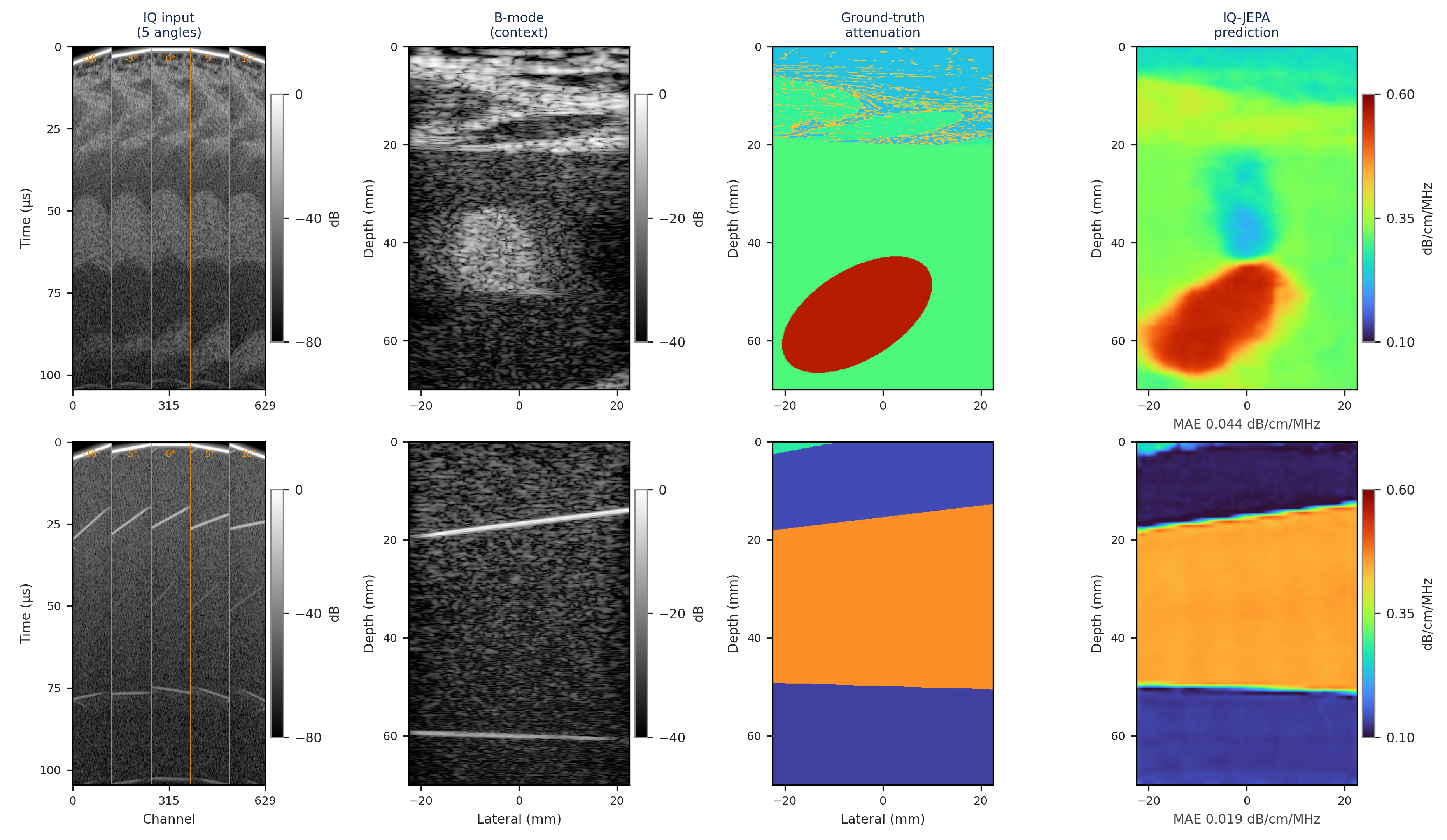}
 \caption{Predictions on two held-out samples, an abdomen with a high-attenuation inclusion (top) and a 4-layer phantom (bottom). Each row shows the IQ channel data, the B-mode reconstruction, the ground-truth attenuation, and the prediction, with per-sample MAE.}
 \label{fig:attenuation-qual}
 \end{subfigure}
 \caption{Attenuation as a second property from the same self-supervised encoder, fine-tuned on attenuation with no new pretraining. Panel (a) is label efficiency, and panel (b) is qualitative predictions from the 63{,}435-label model, which recovers the region-level attenuation and the inclusion.}
 \label{fig:attenuation-combined}
\end{figure}

We test whether the same self-supervised encoder transfers to attenuation, a distinct acoustic property, with no new pretraining. The encoder pretrained with IQ-JEPA is fine-tuned on attenuation labels from the same initialization used for sound speed, and only the fine-tuning regression target differs. Attenuation is reported as a test MAE in dB/cm/MHz against a mean-predictor floor of 0.118. The dataset draws per-pixel material properties independently by region (Section~\ref{sec:methods:dataset}). Sound speed and attenuation are therefore nearly uncorrelated at a correlation coefficient of 0.16, making attenuation a physically distinct target rather than a sound-speed proxy. This independence by design is why we assign the properties at random. A frozen convolutional probe reads attenuation from the self-supervised features at 0.093\,dB/cm/MHz. The same probe on a randomly initialized encoder reaches 0.124\,dB/cm/MHz, which does not beat the mean-predictor floor. A convolutional head therefore cannot recover attenuation from random features. The accessible attenuation structure is a property of the representation rather than the head, mirroring the sound-speed frozen-probe result above. Fine-tuning this encoder then beats matched supervised training at every label count (Figure~\ref{fig:attenuation}). At 10{,}000 labels it reaches 0.0552 against 0.0699\,dB/cm/MHz for supervised training from scratch. Self-supervision lowers the error by 14 to 24\% across the label range. The margin is largest at intermediate label counts and narrows to 14\% at the full 63{,}435-label set as the two converge. A single self-supervised encoder therefore reproduces the sound-speed label-efficiency behavior on a second physically distinct property. This is what the architecture anticipates. Sound speed is a travel-time cue carried by the phase difference in the channel data. Attenuation is a frequency-dependent amplitude decay carried by the magnitude. Both are invariant to the global demodulation phase. The predicted maps recover the region-level attenuation and the localized inclusions of the ground truth (Figure~\ref{fig:attenuation-qual}), with the residual error concentrating at region boundaries. These attenuation runs are deterministic single-seed.

\subsubsection{Cross-distribution transfer}
\label{sec:results:crossdist}
We next test whether the learned representation is tied to one phantom distribution. We pretrain the Hermitian ViT on one in-silico family, abdomen or layered, then fine-tune and evaluate on a target family. The pretraining set is fixed at 20{,}000 acquisitions, so only the pretraining distribution differs between an in-distribution and a cross-distribution condition, not its size (Table~\ref{tab:crossdist}). Pretraining on the other family costs little. The transfer gap, the cross-distribution minus the in-distribution error, is $+0.36$\ms{} on abdomen and $+0.24$\ms{} on layered, both about 2\% of the error and within the seed spread of comparable cells. Both cross-distribution models beat supervised training from scratch at matched labels (25.69\ms{} on abdomen, 19.10\ms{} on layered). So even wrong-family pretraining keeps most of the gain, 4.9 and 3.9\ms{} over supervised. Together with the transfer to attenuation above, this indicates the self-supervised features are largely shared across these families rather than family-specific. These results are within the in-silico Fullwave families. Transfer to real acquisitions is left to future work.

\begin{table}[tb]
 \caption{Cross-distribution transfer for the Hermitian ViT. Held-out test MAE (m/s) at 10{,}000 labels, pretraining fixed at 20{,}000 acquisitions so only its distribution differs. Diagonal (bold) is in-distribution, off-diagonal cross-distribution. The transfer gap (bottom row) is under 2\% of the error, and both cross-distribution models stay below the supervised floor. Single-seed.}
 \label{tab:crossdist}
 \centering
 \small
 \begin{tabular}{lcc}
 \toprule
 & \multicolumn{2}{c}{Evaluation family (test MAE, m/s)} \\
 \cmidrule(lr){2-3}
 Pretraining & abdomen & layered \\
 \midrule
 abdomen & \textbf{20.40} & 15.19 \\
 layered & 20.76 & \textbf{14.95} \\
 supervised (no pretraining) & 25.69 & 19.10 \\
 \midrule
 transfer gap (cross $-$ in) & $+0.36$ & $+0.24$ \\
 \bottomrule
 \end{tabular}
\end{table}

\subsection{Why latent prediction and the complex encoder}
\label{sec:results:ssl-design}

\begin{figure}[tb]
 \centering
 \begin{subfigure}[c]{0.43\textwidth}
 \centering
 \begingroup
 \renewcommand{\tiny}{\fontsize{6.5}{7.8}\selectfont}%
 \renewcommand{\scriptsize}{\fontsize{9.1}{10.9}\selectfont}%
 \renewcommand{\footnotesize}{\fontsize{10.4}{12.5}\selectfont}%
 \renewcommand{\small}{\fontsize{11.7}{14}\selectfont}%
 \renewcommand{\normalsize}{\fontsize{13}{15.6}\selectfont}%
 \adjustbox{max width=\linewidth}{\begin{tikzpicture}
  \begin{axis}[
      iqaxis,
      width=9cm, height=7cm,
      xmode=log, log basis x=10,
      log ticks with fixed point,
      xminorticks=false,
      xtick={1000,5000,10000,63435},
      xticklabels={1K,5K,10K,63K},
      enlarge x limits=0.06,
      xlabel={Label count (log)},
      ylabel={test MAE (m/s)},
      ymin=6, ymax=50,
      legend style={font=\scriptsize, draw=black!25, fill=white,
          at={(0.5,-0.27)}, anchor=north, legend columns=1, cells={anchor=west}},
      reverse legend,
    ]
    \addplot[iqred, mark=*, error bars/.cd, y dir=both, y explicit,
        error bar style={line width=0.9pt}, error mark options={mark size=3.5pt, line width=0.9pt}] coordinates
      {(1000,30.47) +- (0,0.91) (5000,19.72) +- (0,0.35) (10000,15.60) +- (0,0.26) (63435,8.71) +- (0,0.17)};
    \addlegendentry{IQ-JEPA SSL+ft}
    \addplot[iqgreen, mark=square*] coordinates
      {(1000,37.69)(5000,26.59)(10000,21.29)(63435,10.46)};
    \addlegendentry{Masked AE SSL+ft}
    \addplot[iqgray, mark=o, densely dashed, error bars/.cd, y dir=both, y explicit,
        error bar style={line width=0.9pt, solid}, error mark options={mark size=3.5pt, line width=0.9pt, solid}] coordinates
      {(1000,45.44) +- (0,0.72) (5000,29.87) +- (0,1.24) (10000,23.46) +- (0,0.18) (63435,11.51) +- (0,0.15)};
    \addlegendentry{Supervised}
  \end{axis}
\end{tikzpicture}}%
 \endgroup

 \caption{Sound-speed test MAE versus label count. Latent prediction (IQ-JEPA) versus masked-autoencoder pixel reconstruction on the same Hermitian ViT encoder, with the supervised Hermitian ViT as a dashed grey reference. The IQ-JEPA and supervised series are three-seed means with standard-deviation error bars. The masked-autoencoder series is single-seed.}
 \label{fig:ssl-objective-quant}
 \end{subfigure}
 \hfill
 \begin{subfigure}[c]{0.46\textwidth}
 \centering
 \includegraphics[width=1.0\linewidth]{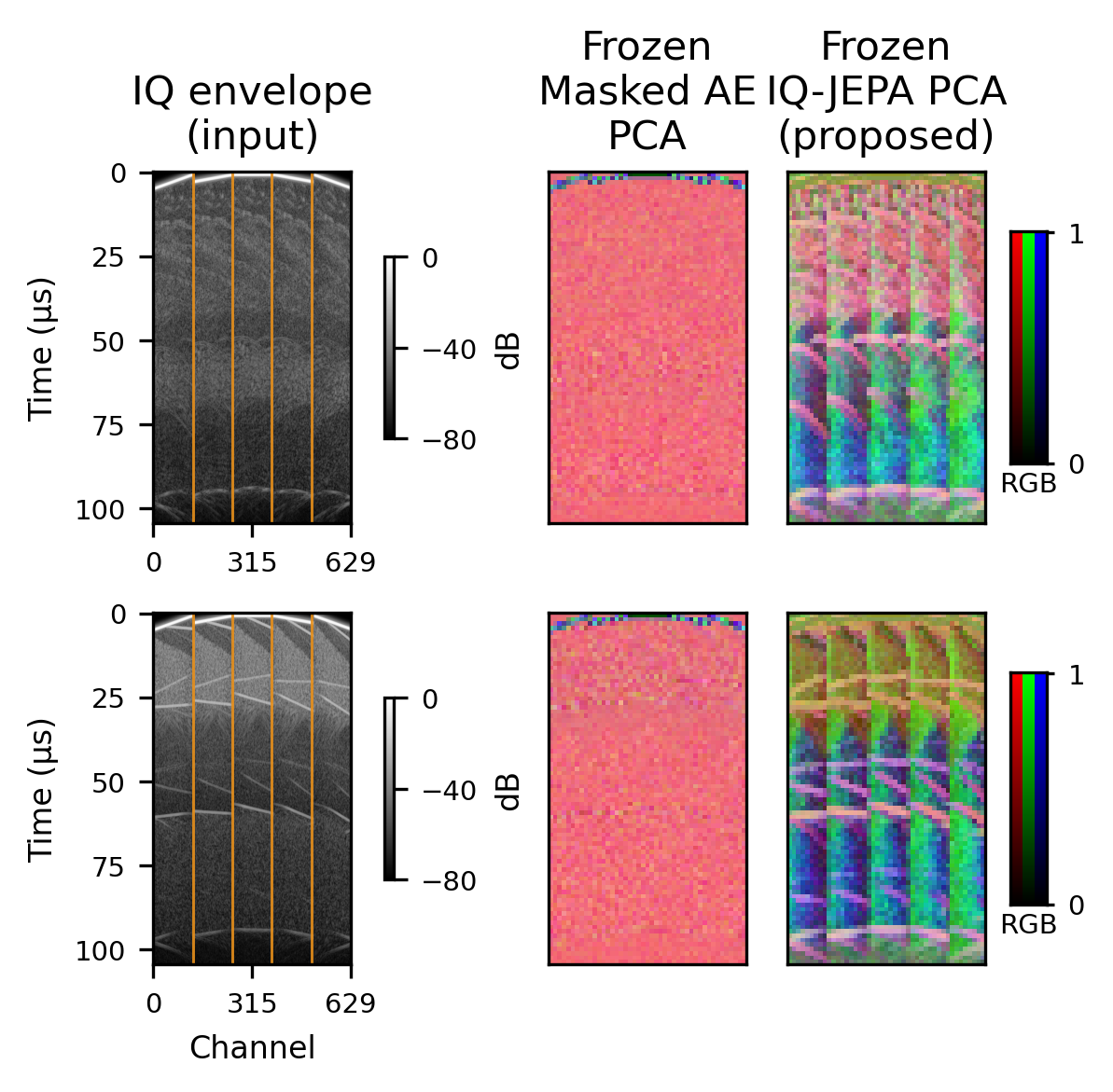}
 \caption{Per-patch features from the two frozen encoders for two held-out samples (rows). Features are projected to RGB by their top three principal components with no labels. IQ-JEPA forms richly structured features, while the masked-autoencoder features stay nearly uniform.}
 \label{fig:learned-features}
 \end{subfigure}
 \caption{Latent prediction versus pixel reconstruction, shown two ways on the same Hermitian ViT encoder, with only the self-supervised objective swapped. (a)~The label-efficiency gap and (b)~the frozen learned features both favor latent prediction.}
 \label{fig:ssl-objective}
\end{figure}

The choice of self-supervised objective matters. The same advantage shows up both quantitatively and qualitatively (Figure~\ref{fig:ssl-objective}). Holding the Hermitian ViT encoder fixed and swapping only the objective, latent prediction (IQ-JEPA) reaches $30.47 / 19.72 / 15.60 / 8.71\ms$ at $1$K$/5$K$/10$K$/63$K labels. A masked-autoencoder pixel-reconstruction objective \citep{He2022-iq} reaches $37.69 / 26.59 / 21.29 / 10.46\ms$ at the same label counts. The latent-prediction advantage of $7.22 / 6.87 / 5.69 / 1.75\ms$ widens as labels grow scarce (Figure~\ref{fig:ssl-objective-quant}). Both objectives improve on supervised training from scratch, and even pixel reconstruction edges it at the largest label count. Latent prediction is the stronger of the two at every label count.

The learned features tell the same story. We project each frozen encoder's per-patch features to RGB by their top three principal components \citep{bardes2024revisiting, MurLabadia2026-vjepa}, on held-out samples and with no labels. This separates the two objectives just as sharply (Figure~\ref{fig:learned-features}). The IQ-JEPA features are richly structured and show acquisition-domain patterns, including the wavefront moveout and the vertical seams between the plane-wave angles. The masked-autoencoder uses the same architecture, data, and masking, yet its features are far weaker and nearly uniform. So the objective, not the architecture, decides whether the frozen features carry any structure. Both readouts point to the same cause. Pixel reconstruction is poorly suited to speckle-dominated IQ, whose exact realization is challenging to predict from a masked context.

\subsection{Which encoder components matter}
\label{sec:results:ablation}
The complex encoder is the other half of the design. To locate where its advantage lies, we run the full $2^5$ factorial over the five factors of the Hermitian ViT (the complex attention, the conjugate-product feed-forward, the complex patch embedding, rotary position encoding, and phase augmentation). The runs are supervised at 10{,}000 labels and deterministic. The 32 configurations span 23.42 to 32.10\ms{} (Appendix~\ref{sec:appendix-factorial}, Table~\ref{tab:factorial}). The full configuration, all five factors on, is the lowest at 23.42\ms, while turning all five off gives 31.66\ms. The matched-real corner of this factorial, with rotary encoding and phase augmentation but no complex components, reaches 24.32\ms, close to the full Hermitian configuration. The supervised effect of the complex block is therefore this small, configuration-gated difference, not the larger gap over the standard real-valued ViT baseline of Table~\ref{tab:label-efficiency} (30.04\ms), which is a different architecture. That baseline is the same real-valued ViT baseline, run from the published implementation \citep{kaggle-waveform-inversion, Sheoran2025-1stplace}. We run the factorial in the supervised setting, which avoids a separate pretraining run for each of the 32 configurations. The single complex block is re-examined under self-supervision across pretraining size in Section~\ref{sec:results:pretrain-scaling}.

\begin{table}[tb]
 \centering
 \caption{Per-factor main effects of the five Hermitian ViT, supervised at 10{,}000 labels. Each value is the change in sound-speed MAE (m/s) when a factor is turned on. Negative is better. ``Cube average'' is the mean over the other factors' $2^4$ settings. ``At full config'' flips one factor from all-on. From the full $2^5$ factorial (Table~\ref{tab:factorial}).}
 \label{tab:main-effects}
 \small
 \begin{tabular}{lrr}
 \toprule
 Factor & Cube average (m/s) & \makecell{At full \\config (m/s)} \\
 \midrule
 Phase augmentation & $-3.10$ & $-6.35$ \\
 Rotary position encoding & $-3.51$ & $-3.69$ \\
 Complex patch embedding & $-1.38$ & $-2.04$ \\
 \makecell[l]{Conjugate-product \\feed-forward} & $+0.48$ & $-0.23$ \\
 Hermitian attention & $+0.44$ & $-0.31$ \\
 \bottomrule
 \end{tabular}
\end{table}

The per-factor main effects separate into two groups (Table~\ref{tab:main-effects}). Three factors are robust. Phase augmentation ($-3.10\ms$ averaged over the cube), rotary position encoding ($-3.51$), and the complex patch embedding ($-1.38$) lower error throughout. Their effect is even larger at the full configuration ($-6.35$, $-3.69$, $-2.04$). The two complex-block factors are configuration gated. The conjugate-product feed-forward and the Hermitian attention each lower error at the full configuration ($-0.23$ and $-0.31$), but their cube-averaged effect flips sign ($+0.48$ and $+0.44$). Away from the full configuration, the unconstrained real alternatives carry twice the parameters and spend the extra capacity to win. We keep them for the exact $U(1)$ symmetry of the task, carried at half the parameters of the unconstrained real alternatives, and because their benefit appears under self-supervision rather than in supervised average-case accuracy. This supervised factorial understates the complex block, whose value appears under self-supervision. There the single complex block is a data-efficiency factor, lowering error by up to 3.1\ms{} at a small pretraining set (Section~\ref{sec:results:pretrain-scaling}), about six times its combined supervised full-config effect of 0.54\ms{}. This is because it contributes to representation learning from unlabeled data rather than to supervised average-case accuracy.

The conjugate-product feed-forward and phase augmentation are complementary rather than substitutes. At the full configuration, the feed-forward lowers MAE by 0.23\ms{} when phase augmentation is on (23.65 to 23.42\ms{}), but raises it by 4.01\ms{} when augmentation is off (25.76 to 29.77\ms{}, Table~\ref{tab:factorial}). The architectural invariance therefore pays off only when training exercises the phase symmetry, so the two mechanisms reinforce rather than replace each other.

Label-free probes confirm that the two complex components carry the $U(1)$ symmetry by construction, invariant and equivariant as required (Appendix~\ref{sec:appendix-symmetry}). %
\section{Discussion}
\label{sec:discussion}

\subsection{Self-supervision yields a transferable encoder}

The clearest pattern across Section~\ref{sec:results} is that self-supervised pretraining moves the sound-speed accuracy more than any other single design choice. Adding IQ-JEPA pretraining to the Hermitian ViT lowers error by 14.97, 10.15, 7.86, and 2.80\ms{} at 1{,}000, 5{,}000, 10{,}000, and 63{,}435 labels. The gains are largest in the low-label regime, and even at the full labeled set the 2.80\ms{} gain is non-trivial. The pretraining-data scaling trend reinforces this from a second angle. Holding labels fixed at 10{,}000 and growing only the unlabeled data lowers error from 23.46 to 15.60\ms. The curve is still descending at the largest set we could assemble. The ratio of pretraining to fine-tuning data here is about 6 to 1 at the full training set, far below the ratios common in self-supervised recipes for language and vision \citep{Devlin2019-bert, Brown2020-gpt3, He2022-iq}. The reported accuracy is therefore a conservative reading of the method rather than its ceiling. The most direct way to push accuracy further is a larger unlabeled set, whether additional simulation or, eventually, real acquisitions. Both the labeled and unlabeled data here are simulated, so this trend measures the value of pretraining data rather than a cost saving. The cost asymmetry it targets, abundant unlabeled acquisitions against scarce simulated labels, is realized once the unlabeled set is drawn from in vivo or ex vivo acquisitions.

Beyond accuracy, the pretrained representation transfers (Section~\ref{sec:results:transfer}). The same frozen encoder reads out both sound speed and attenuation above a random-encoder control. It fine-tunes to attenuation with a smaller but consistent label-efficiency gain. It also largely carries across phantom distributions, with a small cross-distribution penalty. So the same self-supervised encoder transfers to multiple downstream properties and distributions rather than a single task. This reuse of one encoder across several tasks is the kind of transfer that motivates foundation models, shown here at the scale of an in-silico proof of concept.

\subsection{Complex IQ and latent prediction}

Complex-valued IQ and latent prediction fit the same physics. Travel time is the primary cue for sound speed. In the IQ signal, it appears as a phase difference that is invariant to the global demodulation phase. The Hermitian ViT preserves that phase difference by construction (Section~\ref{sec:methods:encoder}), and classical coherence methods exploit an analogous quantity \citep{Jaeger2015-dq}. Latent prediction is the matching objective. Pixel reconstruction is the poorer fit, consistent with the 5.69\ms{} advantage of latent prediction over masked-autoencoder reconstruction at 10{,}000 labels (Section~\ref{sec:results:ssl-design}).

The complex encoder helps most where labels are scarcest. With matched IQ-JEPA pretraining, the Hermitian ViT improves on a real-valued ViT by 3.46\ms{} at 1{,}000 labels. The gap narrows to 1.28\ms{} at 63{,}435 but does not close. Both conditions are three-seed means, and the margin exceeds the seed spread at every label count. The complex inductive bias is most valuable when the model has too few examples to learn the phase structure implicitly. This is the label-scarce regime relevant to clinical translation, where labeled ground truth is the binding constraint. The Hermitian flash-attention construction (Appendix~\ref{sec:appendix-flash}) makes this practical. It brings the cost of the complex front end to about $1.7\times$ a real layer, whereas explicit complex matrix multiplication costs $4$ to $8\times$. As a property of the Hermitian inner product, it transfers to any complex-valued transformer.

\subsection{Channel data versus beamformed images}

A natural alternative is to estimate sound speed from the reconstructed B-mode image rather than from the raw channel data. We use the channel data deliberately. Forming a B-mode image requires beamforming at a fixed assumed sound speed, which writes a constant-speed bias into the image geometry. It displaces scatterers from their true positions by the very quantity we are trying to recover. The delay-and-sum step that builds the image then discards the per-channel phase, collapsing the coherent wavefront into a single envelope sample per pixel. That per-channel phase also records the wavefront aberration that sound-speed heterogeneity imposes. This aberration is substantial in tissue such as the abdominal wall \citep{Hinkelman1994-aberration}. Raw RF and IQ data retain both the undistorted travel-time geometry and the channel phase. They therefore carry more information for sound-speed estimation than the B-mode image derived from them. For anatomical segmentation, a B-mode image is the natural and desirable input. Quantitative sound-speed estimation needs the exact scatterer geometry free of the beamforming speed bias, which only the channel data preserve. A model pretrained on channel data also learns the coherent wavefront structure that beamforming itself relies on. A model trained on B-mode images no longer has access to the per-channel phase that beamforming discards. This argument is physical, and we have not yet tested it in this paper. The comparison against a B-mode encoder remains future work.

\subsection{Relation to prior work}

Pulse-echo sound-speed estimation and seismic full-waveform inversion are a similar wave-speed inverse problem in different regimes. Ultrasound works at megahertz rather than the hertz to kilohertz frequencies of seismic exploration. It uses a compact pulse-echo aperture rather than a wide-offset transmission survey, and it faces speckle and a low signal-to-noise ratio that the seismic setting does not. Full-waveform inversion \citep{Pratt1999-wc} and coherence-based estimation \citep{Jaeger2015-dq} solve this problem iteratively and use no training data. The learned solvers, InversionNet \citep{Wu2019-inversionnet} and the OpenFWI transformers \citep{Deng2022-qo, kaggle-waveform-inversion, Sheoran2025-1stplace}, run in milliseconds but are label hungry. The fine-tuned encoder keeps the fast feed-forward inference of a learned solver, while IQ-JEPA pretraining addresses the label cost through self-supervision. It differs from the OpenFWI lineage in two ways. It operates on native complex-valued IQ with a complex encoder rather than on real-valued inputs with a real network, and it pretrains with latent prediction. One benchmark shows the phase of the IQ signal stabilizes learned recovery against operator-dependent effects \citep{Feigin2024-sos}, which our phase-aware design exploits from the input onward. Learned sound-speed recovery from raw channel data has also used autoencoder architectures, including linked autoencoders on pulse-echo raw data \citep{Jush2023-fc} and networks comparing coherency-based against RF-data inputs \citep{Heller2023-fi}. IQ-JEPA differs in two ways. It predicts the latent representation of masked regions rather than reconstructing the raw signal, which our own comparison finds stronger than pixel reconstruction (Section~\ref{sec:results:ssl-design}). And it operates on the complex IQ with a phase-preserving encoder rather than on real-valued inputs.

Complex-valued networks are an established line, well suited to wave signals where phase encodes travel time \citep{Trabelsi2018-complex, Lee2022-cvnnsurvey}. Their advantage is conditional on the task respecting the relevant phase symmetry rather than universal \citep{Kumar2026-cvnnhelp}. This is consistent with our finding that the complex blocks earn their place on symmetry grounds rather than as average-case accuracy factors. Complex-valued transformers have been proposed \citep{Yang2020-complextransformer, Eilers2023-blocks}, and an efficient flash-attention realization is available in open-source code \citep{Wang2023-cvt}. Our Hermitian attention coincides with the strongest complex-attention block of \citet{Eilers2023-blocks}. Our contribution there is its structure-preserving use for the phase-invariant ultrasound task. In ultrasound specifically, complex-valued networks have been applied directly to IQ or analytic channel data for beamforming \citep{Zhang2025-ccgr}, fast reconstruction \citep{Lu2022-cidnet, Bentaleb2024-cresattunet}, clutter filtering \citep{Han2025-clunet}, and aberration correction \citep{Xing2024-cvcnn}. These are supervised and target image formation. Self-supervised learning has separately been applied to ultrasound RF, for image reconstruction \citep{Zhang2021-pwus} and denoising \citep{Huang2026-ha2ha}, where the supervision comes from physical or angular-consistency losses rather than a transferable encoder. IQ-JEPA sits between these two lines. It uses the complex representation for self-supervised representation learning of a quantitative inverse problem, predicts latent representations of masked IQ rather than reconstructing the signal, and yields one encoder that transfers across properties.

The closest architectures to ours are recent complex-valued ViTs for image classification. One operates on raw complex MRI k-space \citep{Rempe2026-kvit}, and the other converts real images to the complex domain \citep{Sabokpa2026-cvt}. Both are built from a complex patch embedding, complex positional encoding, and complex attention. We differ in three ways. The first is the signal and task. We use pulse-echo IQ channel data for quantitative sound-speed regression rather than image classification. The second is the self-supervised IQ-JEPA objective, which neither uses. The third is the conjugate-product feed-forward, which computes a phase difference analogous to the one coherence methods use for sound speed. Against the convolutional baseline, IQ-JEPA is about $2.2\times$ lower at 10{,}000 labels (15.60 versus 34.73\ms). This is consistent with global attention over the token grid indexed by depth and angle being better matched to the long-range travel-time cues than a local convolutional receptive field.

\subsection{Limitations}

This study is entirely in silico, a deliberate first step. It remains a proof of concept at a single transmit mode (5-angle plane wave) and a single center frequency (2.5\,MHz), and it has not yet been applied to real data. The two complex-block components are configuration gated. They help at the full configuration and on symmetry grounds, but their cube-averaged effect is neutral to slightly negative. The cross-distribution transfer is within the in-silico Fullwave families. Generalization across transmit geometries and frequencies, broader multi-property transfer, and validation on real acquisitions remain future work.

\subsection{Future work}

The path from this proof of concept to a useful encoder runs through three steps. The first is real data. The pipeline admits in vivo and ex vivo acquisitions into the same unlabeled pretraining set. The natural deployment path is therefore to pretrain on abundant real IQ and fine-tune with in silico labels, which narrows the simulation to real gap. The second is generalization across acquisition. The question is whether an encoder pretrained on plane-wave IQ transfers to different transmit sequences and to higher center frequencies such as 5\,MHz, with or without further pretraining. The third is broader multi-property transfer. Fine-tuning the same self-supervised encoder already recovers attenuation as well as sound speed (Section~\ref{sec:results:attenuation}). Density is a natural next property but is more weakly identified from pulse-echo data. There boundaries return acoustic-impedance products rather than density alone, so it needs a dedicated clean baseline. Beyond quantitative property mapping, the encoder's learned representation of the channel data could also serve image reconstruction tasks such as adaptive beamforming, aberration correction, and reverberation correction. These all depend on the coherent structure of the channel data.

\subsection{Broader applicability}

The recipe of complex IQ input, a complex-valued vision transformer, and a latent-prediction objective is not specific to ultrasound. Wherever a coherent imaging modality records a complex field and the inverse problem is label limited, such as seismic full-waveform inversion, synthetic-aperture radar, and magnetic resonance imaging, the same framework may apply.
\section{Conclusion}
\label{sec:conclusion}

We introduced IQ-JEPA, a self-supervised method for quantitative ultrasound that operates directly on complex-valued IQ channel data. The encoder is a Hermitian Vision Transformer trained with a latent-prediction objective. On Fullwave 2.5 plane-wave simulations at 2.5\,MHz, self-supervised pretraining lowers sound-speed error from 23.46 to 15.60\ms{} at 10{,}000 labels. This is a roughly threefold gain in label efficiency over the matched supervised model, growing to more than fourfold at 1{,}000 labels and about $2.2\times$ lower than an InversionNet baseline (15.60 versus 34.73\ms). At the largest label count, error reaches 8.71\ms{}. Three findings support the approach. Self-supervision is the dominant factor, larger than any architectural choice we tested. Downstream error follows a pretraining-data scaling trend that is still descending at the largest pretraining set. Additional unlabeled IQ is therefore the most direct route to further gains. At matched probe capacity, a frozen encoder already reads out sound speed far better than a random one (45.64 versus 70.29\ms), and its best frozen readout reaches 22.34\ms{} against the 15.60\ms{} fine-tuned model, so the features already encode sound speed learned without labels. Fine-tuned on attenuation, the same encoder recovers a second acoustic property with a smaller but consistent label-efficiency gain. The representation also largely transfers across phantom distributions, showing it is not tied to one property or one distribution. The architecture is justified on symmetry grounds. The conjugate-product feed-forward is invariant to the global demodulation phase, and the Hermitian attention is equivariant to it. Together they are the exact $U(1)$ structure that a phase-invariant target requires. These results are an in-silico proof of concept. Extending the same pipeline across transmit geometries, center frequencies, and real in vivo and ex vivo acquisitions is the path toward a clinically useful self-supervised encoder for quantitative ultrasound.
 
\section*{Acknowledgments}
The authors thank the University of North Carolina at Chapel Hill and the Research Computing group, in particular Rob Zelt, for computational resources and support. The authors also thank their colleagues for fruitful discussions.

\bibliographystyle{IEEEtranN}
\bibliography{references}

\appendices
\counterwithin{figure}{section}
\counterwithin{table}{section}

\section{Implementation details}
\label{sec:appendix-impl}

Table~\ref{tab:hyperparameters} lists the training and architecture hyperparameters. The unlabeled pretraining set is up to 63{,}435 acquisitions (the full training set) for the headline label-efficiency and pretraining-scaling results. The cross-distribution transfer cells share a pretraining set fixed at 20{,}000 acquisitions. The supervised $2^5$ ablation is at 10{,}000 labels. The headline Hermitian ViT IQ-JEPA SSL+ft results (Table~\ref{tab:label-efficiency}, Figure~\ref{fig:scaling}a) are the mean over three seeds. Each seed is a full pretraining run and fine-tuning ladder. Every estimator of Table~\ref{tab:label-efficiency} is a three-seed mean over seeds 42, 123, and 7. The three-seed spread is small and shrinks as labels grow, from 0.91\ms{} at 1{,}000 labels to 0.26\ms{} at 10{,}000 and 0.17\ms{} at the full label count, well inside the margin over supervised training. We report the mean rather than the best of the three seeds at every label count, so the headline is conservative. The predictor and the exponential moving average target follow the standard I-JEPA design \citep{Assran2023-ijepa}. The regression head is a lightweight batch-normalized convolutional stack mapping the encoder's final tokens to the sound-speed grid. The masked-autoencoder baseline (Section~\ref{sec:results:ssl-design}) is matched to IQ-JEPA in everything but the objective. It uses the same encoder, the same four-block masking (target-block scale $0.25$ to $0.35$), and a six-layer transformer decoder of width 384. The reconstruction target is the per-patch-normalized IQ, with the in-phase and quadrature parts predicted jointly as two channels under a mean-squared-error loss on the masked patches. Only the objective, latent prediction versus pixel reconstruction, differs from IQ-JEPA.

\begin{widetable}[tb]
 \caption{Training and architecture hyperparameters.}
 \label{tab:hyperparameters}
 \centering
 \small
 \begin{tabular}{lll}
 \toprule
 & Pretraining (IQ-JEPA) & Fine-tuning \\
 \midrule
 Objective & smooth-$\ell_1$ latent prediction & normalized MSE regression \\
 Optimizer & AdamW & AdamW \\
 Peak learning rate & $1.5 \times 10^{-4}$ & $1.5 \times 10^{-4}$ \\
 Weight decay & 0.05 & 0.05 \\
 Schedule & 10-epoch warmup, cosine to 0 & 5-epoch warmup, cosine \\
 Epochs & 100 & 100 \\
 Batch size & 16 per GPU (128 effective) & 16 per GPU \\
 EMA momentum $\tau$ & $0.996 \rightarrow 1.0$ (linear) & -- \\
 Hardware & \multicolumn{2}{l}{2 nodes $\times$ 4 NVIDIA H100 (80\,GB)} \\
 \midrule
 \multicolumn{3}{l}{\emph{Encoder (Hermitian ViT, ViT-S scale)}} \\
 Patch size & \multicolumn{2}{l}{$14 \times 14$} \\
 Complex embedding dimension & \multicolumn{2}{l}{192 (384 real after concatenation)} \\
 Depth & \multicolumn{2}{l}{12 blocks (1 complex $+$ 11 real)} \\
 Attention heads & \multicolumn{2}{l}{6} \\
 MLP ratio & \multicolumn{2}{l}{4.0} \\
 Parameters & \multicolumn{2}{l}{21.9\,M} \\
 \midrule
 \multicolumn{3}{l}{\emph{Joint-embedding predictive masking}} \\
 Context coverage & \multicolumn{2}{l}{$\approx 28\%$ of patches} \\
 Target regions & \multicolumn{2}{l}{\makecell[l]{4 per sample, 0.25 to 0.35 of patches each \\angle stripe, depth band, lateral stripe, or block}} \\
 \bottomrule
 \end{tabular}
\end{widetable}

\section{Hermitian flash attention}
\label{sec:appendix-flash}

\paragraph{The complex product is the bottleneck} Computing the complex product $Q\,K^{\mathsf{H}}$ directly is what makes complex attention expensive. It rules out the fused, IO-aware kernels that real-valued vision transformers rely on for speed. In our profiling, it costs four to eight times the wall clock of a matched real attention layer. The Hermitian score, however, is real, and that alone removes the need for the complex product. Writing $Q = Q_r + i\,Q_i$ and $K = K_r + i\,K_i$, we have the identity
\begin{equation}
 \mathrm{Re}(Q\,K^{\mathsf{H}}) = Q_r K_r^{\top} + Q_i K_i^{\top} = [Q_r \,\Vert\, Q_i]\,[K_r \,\Vert\, K_i]^{\top}
 \label{eq:hermitian-identity}
\end{equation}

\paragraph{The Hermitian score reduces to real attention} Equation~\eqref{eq:hermitian-identity} shows that the score is the ordinary real dot product of the queries and keys once their real and imaginary parts are stacked along the feature axis. Complex attention therefore reduces to a single real attention on width-$2 d_h$ vectors, which a standard fused kernel evaluates unchanged. Our Hermitian attention coincides with the strongest complex-attention block of \citet{Eilers2023-blocks}. Routing it through a hardware-fused flash-attention kernel \citep{Dao2022-flashattention} follows an open-source implementation \citep{Wang2023-cvt}. We adopt it because it is what makes the encoder practical to pretrain at our token counts. The construction (Algorithm~\ref{alg:hermitian-flash}) is mechanical. We stack the real and imaginary parts of $Q$, $K$, and $V$ along the head-feature axis, then pass them once through the fused kernel and split the output to recover the complex result. Queries and keys are magnitude-normalized per token, as is now standard in vision transformers.

\begin{algorithm}[tb]
 \caption{Hermitian flash attention.}
 \label{alg:hermitian-flash}
 \begin{algorithmic}[1]
 \Require $Q, K, V \in \mathbb{C}^{B \times H \times N \times d_h}$ as real/imaginary parts
 \State $\tilde{Q} \gets [Q_r \,\Vert\, Q_i]$, $\tilde{K} \gets [K_r \,\Vert\, K_i]$, $\tilde{V} \gets [V_r \,\Vert\, V_i]$
 \Statex \hfill $\triangleright$ concatenate on the feature axis, width $2 d_h$
 \State $O \gets \textsc{FlashAttention}(\tilde{Q},\, \tilde{K},\, \tilde{V};\ \mathrm{scale} = d_h^{-1/2})$
 \Statex \hfill $\triangleright$ single fused, IO-aware call
 \State $\mathrm{out}_r \gets O[\dots, {:}d_h]$, $\mathrm{out}_i \gets O[\dots, d_h{:}]$
 \Statex \hfill $\triangleright$ split to recover the real and imaginary parts
 \State \Return $\mathrm{out}_r + i\,\mathrm{out}_i$
 \end{algorithmic}
\end{algorithm}

\paragraph{The reduction is exact, and one fused call suffices} The reduction is exact rather than approximate, evaluating Eq.~\eqref{eq:hermitian-attn} itself. We verified numerically that the fused output and its gradients match the explicit complex computation to floating-point precision. Since $\tilde{Q}\,\tilde{K}^{\top} = \mathrm{Re}(Q\,K^{\mathsf{H}})$ is real, the softmax yields a single real attention matrix $A \in \mathbb{R}^{N \times N}$. Because $A$ mixes along the token axis while the stacking is on the feature axis, the product $A\,\tilde{V} = [A V_r \,\Vert\, A V_i]$ splits cleanly back into the complex output. A genuinely complex score would instead yield a complex attention matrix that no real fused kernel can evaluate. This fusion works only because the Hermitian score is real. The reduction is also efficient because the kernel never materializes the $N \times N$ score. Memory therefore grows linearly in the token count rather than as the $O(N^2)$ of an explicit complex score. The only overhead is doubling the head-feature width to $2 d_h$, a small constant factor in place of four to eight times the cost of the explicit product.

\paragraph{Score scale} After concatenation the fused kernel sees width $2 d_h$ and, left to its default, would scale by $(2 d_h)^{-1/2}$. We instead pass $d_h^{-1/2}$, scoring at the complex head dimension $d_h$ rather than the flattened width $2 d_h$. This is a $\sqrt{2}$ sharpening of the default, which we found over-smooths the scores and slows pretraining. Existing complex-transformer code keeps the flattened-width default \citep{Wang2023-cvt}.

\paragraph{Cost} Table~\ref{tab:flash-benchmark} reports the resulting cost. At the operating point of 3{,}600 tokens (complex embedding dimension 192, six heads), Hermitian flash attention runs at $1.7\times$ the forward cost of a matched real layer. The explicit, non-fused product runs at $6.9\times$. The gap to real attention narrows to $1.2$ to $1.3\times$ at larger token counts, as the attention core comes to dominate the fixed projection overhead. The memory difference is starker than the latency. The fused kernel holds peak memory roughly linear in the token count. The explicit score grows quadratically and exhausts memory at 16{,}384 tokens, where the fused path still fits in under a gigabyte. All of this follows from the Hermitian inner product being real. The same layer therefore accelerates any complex-valued transformer that uses a Hermitian attention score, not only ours.

\begin{table}[tb]
 \centering
 \caption{Hermitian flash attention overhead versus a matched real-valued attention layer, over a token-count sweep at 6 heads and per-head dimension 32, at batch 4 on an NVIDIA GeForce RTX 4090 (bf16). Each ratio is complex latency over real latency at the same size, so $1\times$ matches real attention and lower is better. Flash is the fused Hermitian kernel and expl.\ the explicit non-fused complex score. Peak memory is the forward+backward footprint, ``OOM'' is out of memory, and $^\star$ marks the operating point.}
 \label{tab:flash-benchmark}
 \small
 \begin{tabular}{r rr rr rr}
 \toprule
 & \multicolumn{2}{c}{forward ratio} & \multicolumn{2}{c}{fwd+bwd ratio} & \multicolumn{2}{c}{peak mem (MB)} \\
 \cmidrule(lr){2-3}\cmidrule(lr){4-5}\cmidrule(lr){6-7}
 Tokens & flash & expl. & flash & expl. & flash & expl. \\
 \midrule
 1024 & 2.7$\times$ & 4.8$\times$ & 2.3$\times$ & 3.6$\times$ & 78 & 251 \\
 3600$^\star$ & 1.7$\times$ & 6.9$\times$ & 1.4$\times$ & 5.3$\times$ & 223 & 2613 \\
 8192 & 1.5$\times$ & 8.0$\times$ & 1.2$\times$ & 6.2$\times$ & 477 & 13133 \\
 16384 & 1.3$\times$ & OOM & 1.2$\times$ & OOM & 932 & OOM \\
 \bottomrule
 \end{tabular}
\end{table}

\section{Latent features by encoder architecture}
\label{sec:appendix-latent-architecture}

\begin{figure}[tb]
 \centering
 \includegraphics[width=0.7\linewidth]{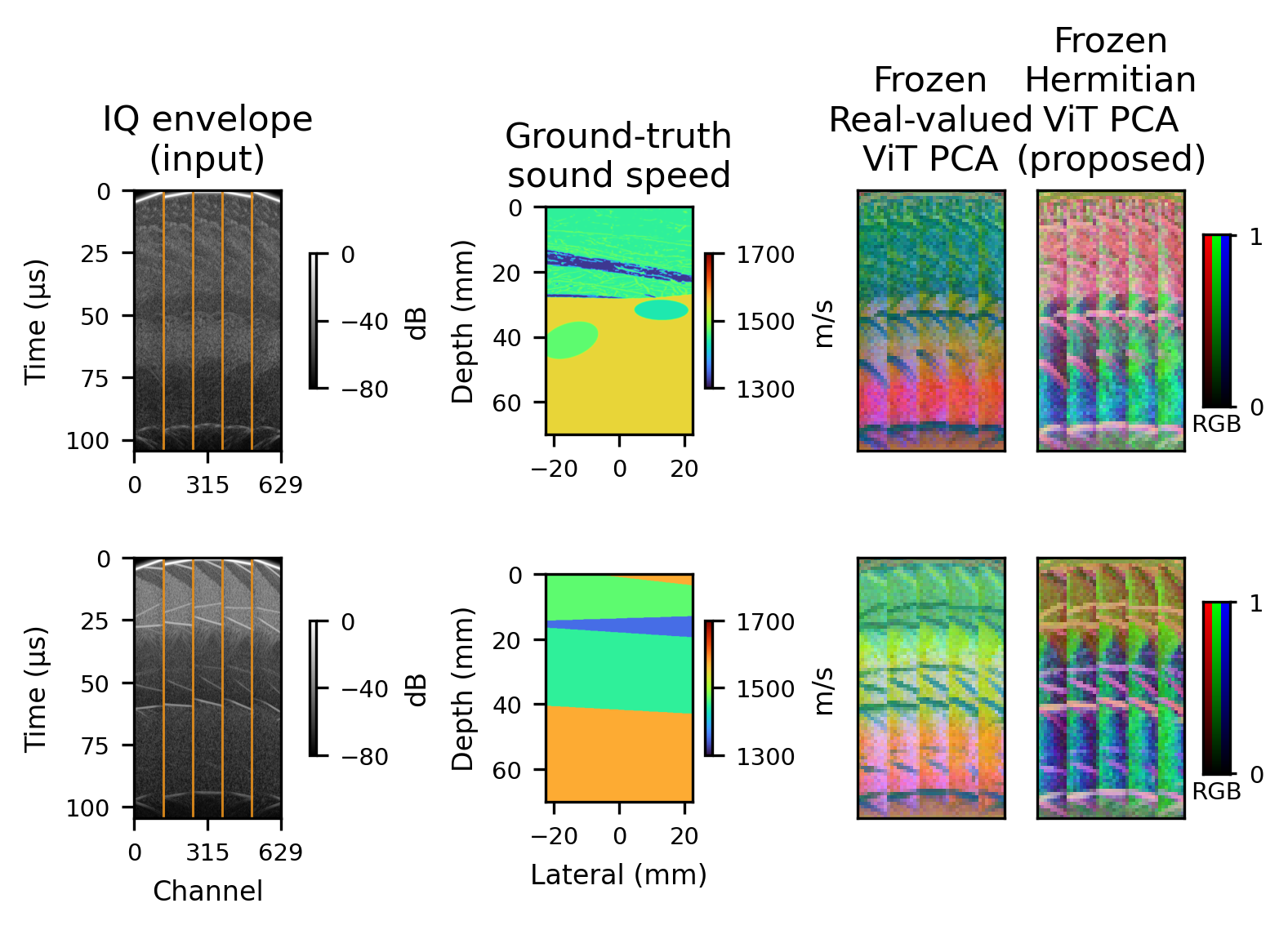}
 \caption{Frozen IQ-JEPA features by encoder architecture, for two held-out samples (rows). Columns are the IQ envelope input, the ground-truth sound speed, and the frozen per-patch features of a real-valued ViT and of the Hermitian ViT, both pretrained with IQ-JEPA on the same dataset and each projected to RGB by its top three principal components (independent basis per column). With no labels, both retain acquisition-domain structure (angle-boundary banding and wavefront moveout) rather than the depth-layered sound-speed map. The Hermitian features additionally show a pronounced vertical striping the real-valued ViT lacks, a qualitative signature of the complex-valued encoder.}
 \label{fig:appendix-latent-architecture}
\end{figure}

\begin{figure}[tb]
 \centering
 \includegraphics[width=0.7\linewidth]{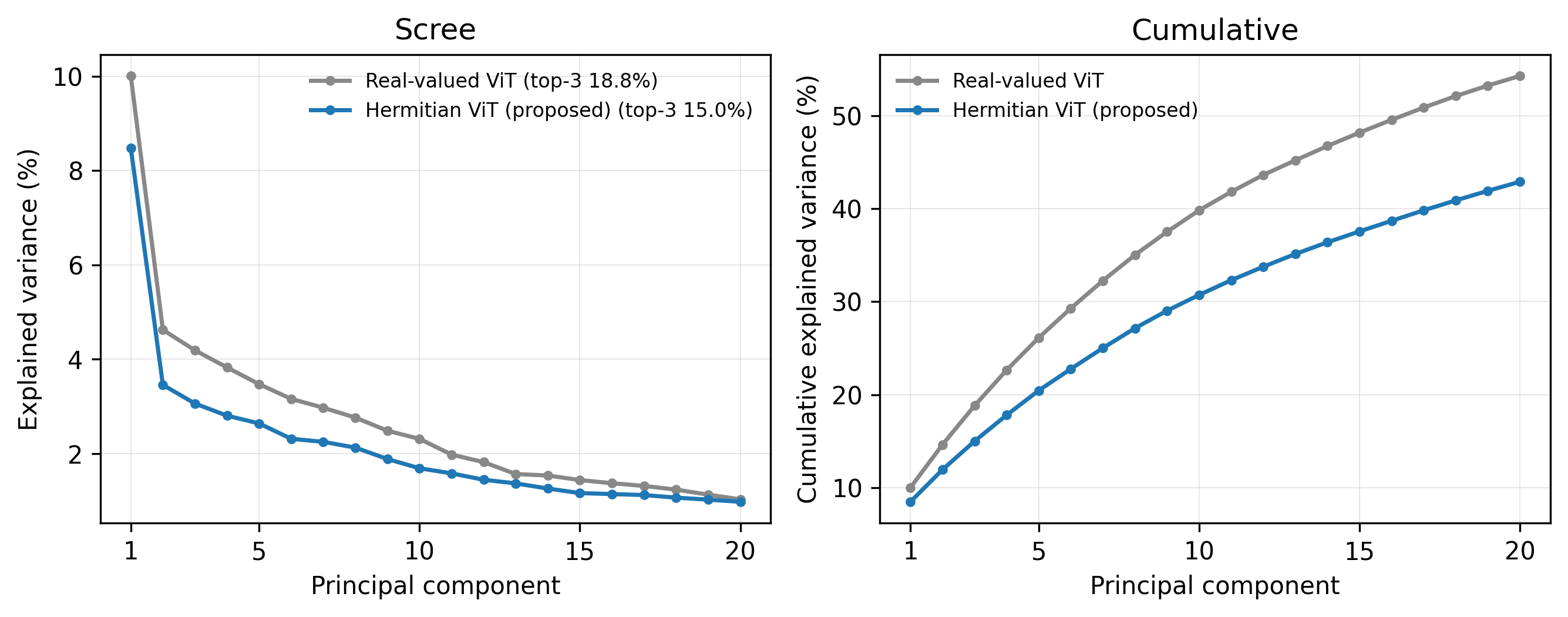}
 \caption{Scree comparison of the frozen IQ-JEPA features by encoder architecture, on the same held-out samples with no labels. The left panel is the per-component explained variance and the right the cumulative, for a real-valued ViT and the Hermitian ViT. The Hermitian spectrum is lower and flatter throughout, spreading its variance across more directions, and its top three principal components capture less of the total (15.0\% versus 18.8\%).}
 \label{fig:appendix-scree}
\end{figure}

Figure~\ref{fig:appendix-latent-architecture} compares the learned representation across encoder architecture. The comparison holds the self-supervised objective and the IQ input fixed and swaps only the encoder. It projects the frozen per-patch features of a real-valued ViT and of the Hermitian ViT to RGB by their top three principal components, with no labels. Both encoders are pretrained with IQ-JEPA on the same dataset and recipe. Both organize in the acquisition domain, showing banding at the concatenated plane-wave angle boundaries and the wavefront moveout across channels rather than the depth-layered sound-speed map. Still, the two encoders differ in a reproducible way. The Hermitian features show a pronounced vertical striping that runs down the travel-time axis. The real-valued ViT features lack this striping and vary more smoothly. The Hermitian variance is also more spread out. Its top three principal components capture 15.0\% of the patch-feature variance, against 18.8\% for the real-valued ViT. The full explained-variance spectrum confirms this, with the Hermitian per-component and cumulative curves lying below the real-valued ViT throughout (Figure~\ref{fig:appendix-scree}). The same striping recurs in the Hermitian features of every sample in Appendix~\ref{sec:appendix-finetune-projection}. We read it as a qualitative signature of the complex-valued encoder. It may reflect the complex-valued processing, but the principal-component view does not establish the mechanism. The quantified effect remains the fine-tuned accuracy margin over the real-valued ViT (Section~\ref{sec:results:label-efficiency}). Fine-tuning then maps these frozen features onto the sound-speed layout (Appendix~\ref{sec:appendix-finetune-projection}).

\section{Full $2^5$ factorial}
\label{sec:appendix-factorial}

\begin{widetable}[tb]
 \centering
 \caption{Full $2^5$ factorial, supervised at 10{,}000 labels. Values are deterministic sound-speed test MAE (m/s). $\bullet$ is the complex or on setting (complex attention, conjugate-product feed-forward, complex patch embedding, RoPE, phase augmentation) and $\circ$ the real or off setting. The two panels are one 32-row table.}
 \label{tab:factorial}
 \scriptsize
 \begin{minipage}{0.43\textwidth}
 \centering
 \setlength{\tabcolsep}{3pt}
 \begin{tabular}{ccccc r}
 \toprule
 \shortstack{Complex \\attention} & \shortstack{Conjugate\\MLP} & \shortstack{Patch\\embedding} & RoPE & \shortstack{Phase\\augmentation} & MAE \\
 \midrule
 $\bullet$ & $\bullet$ & $\bullet$ & $\bullet$ & $\bullet$ & \textbf{23.42} \\
 $\bullet$ & $\bullet$ & $\bullet$ & $\bullet$ & $\circ$ & 29.77 \\
 $\bullet$ & $\bullet$ & $\bullet$ & $\circ$ & $\bullet$ & 27.11 \\
 $\bullet$ & $\bullet$ & $\bullet$ & $\circ$ & $\circ$ & 29.74 \\
 $\bullet$ & $\bullet$ & $\circ$ & $\bullet$ & $\bullet$ & 25.46 \\
 $\bullet$ & $\bullet$ & $\circ$ & $\bullet$ & $\circ$ & 30.80 \\
 $\bullet$ & $\bullet$ & $\circ$ & $\circ$ & $\bullet$ & 28.97 \\
 $\bullet$ & $\bullet$ & $\circ$ & $\circ$ & $\circ$ & 32.10 \\
 $\bullet$ & $\circ$ & $\bullet$ & $\bullet$ & $\bullet$ & 23.65 \\
 $\bullet$ & $\circ$ & $\bullet$ & $\bullet$ & $\circ$ & 25.76 \\
 $\bullet$ & $\circ$ & $\bullet$ & $\circ$ & $\bullet$ & 28.12 \\
 $\bullet$ & $\circ$ & $\bullet$ & $\circ$ & $\circ$ & 30.65 \\
 $\bullet$ & $\circ$ & $\circ$ & $\bullet$ & $\bullet$ & 25.45 \\
 $\bullet$ & $\circ$ & $\circ$ & $\bullet$ & $\circ$ & 27.11 \\
 $\bullet$ & $\circ$ & $\circ$ & $\circ$ & $\bullet$ & 29.04 \\
 $\bullet$ & $\circ$ & $\circ$ & $\circ$ & $\circ$ & 31.73 \\
 \bottomrule
 \end{tabular}
 \end{minipage}\hfill
 \begin{minipage}{0.43\textwidth}
 \centering
 \setlength{\tabcolsep}{3pt}
 \begin{tabular}{ccccc r}
 \toprule
 \shortstack{Complex \\attention} & \shortstack{Conjugate\\MLP} & \shortstack{Patch\\embedding} & RoPE & \shortstack{Phase\\augmentation} & MAE \\
 \midrule
 $\circ$ & $\bullet$ & $\bullet$ & $\bullet$ & $\bullet$ & 23.73 \\
 $\circ$ & $\bullet$ & $\bullet$ & $\bullet$ & $\circ$ & 28.11 \\
 $\circ$ & $\bullet$ & $\bullet$ & $\circ$ & $\bullet$ & 27.24 \\
 $\circ$ & $\bullet$ & $\bullet$ & $\circ$ & $\circ$ & 29.14 \\
 $\circ$ & $\bullet$ & $\circ$ & $\bullet$ & $\bullet$ & 24.86 \\
 $\circ$ & $\bullet$ & $\circ$ & $\bullet$ & $\circ$ & 28.45 \\
 $\circ$ & $\bullet$ & $\circ$ & $\circ$ & $\bullet$ & 28.56 \\
 $\circ$ & $\bullet$ & $\circ$ & $\circ$ & $\circ$ & 31.71 \\
 $\circ$ & $\circ$ & $\bullet$ & $\bullet$ & $\bullet$ & 23.96 \\
 $\circ$ & $\circ$ & $\bullet$ & $\bullet$ & $\circ$ & 25.08 \\
 $\circ$ & $\circ$ & $\bullet$ & $\circ$ & $\bullet$ & 27.83 \\
 $\circ$ & $\circ$ & $\bullet$ & $\circ$ & $\circ$ & 30.98 \\
 $\circ$ & $\circ$ & $\circ$ & $\bullet$ & $\bullet$ & 24.32 \\
 $\circ$ & $\circ$ & $\circ$ & $\bullet$ & $\circ$ & 27.34 \\
 $\circ$ & $\circ$ & $\circ$ & $\circ$ & $\bullet$ & 28.85 \\
 $\circ$ & $\circ$ & $\circ$ & $\circ$ & $\circ$ & 31.66 \\
 \bottomrule
 \end{tabular}
 \end{minipage}
\end{widetable}

Table~\ref{tab:factorial} lists the full $2^5$ factorial over the five Hermitian ViT factors, supervised at 10{,}000 labels. The per-factor main effects of Table~\ref{tab:main-effects} are computed from it. The full configuration with all five factors on is the lowest at 23.42\ms{}, while turning all five off gives 31.66\ms.

\section{InversionNet baseline}
\label{sec:appendix-inversionnet}

InversionNet \citep{Wu2019-inversionnet} is run on RF input and reaches $34.73\pm0.07$\ms{} at 10{,}000 labels, a three-seed mean over seeds 42, 123, and 7. The IQ-JEPA Hermitian ViT is about $2.2\times$ lower (15.60\ms). The baseline follows the architecture of the original InversionNet \citep{Wu2019-inversionnet}, with the input adapted to the per-element RF channel data. We did not inherit its learning rate. The original setting trains poorly on this data, reaching only about 70 to 75\ms{} across the label range, so we swept the learning rate and selected $10^{-3}$, which is the configuration reported everywhere in this paper. The baseline is therefore tuned in its own favor rather than inherited.

\section{Frozen-probe head capacity}
\label{sec:appendix-probe}

As a complementary probe of feature accessibility, we freeze the self-supervised encoder and train only a probe head of increasing capacity (linear, convolutional, three-layer convolutional, dilated, and attention). The frozen readout improves monotonically with head capacity, from $62.05\ms{}$ with a linear head to $22.34\ms{}$ with an attention head. Even the highest-capacity head stays well above the $15.60\ms{}$ fine-tuned model. A frozen encoder with a small head cannot fuse the multi-angle channels into the image domain (Figure~\ref{fig:appendix-probe-ladder}). A randomly initialized encoder is the decoder-capacity control. The self-supervised features read sound speed better than random at every head capacity, $62.05$ versus $74.53$, $45.64$ versus $70.29$, $36.74$ versus $63.70$, $27.37$ versus $48.37$, and $22.34$ versus $36.22\ms{}$ from the linear to the attention head. The readout therefore reflects the representation, not the probe head. The advantage traces an inverted-U over the capacity axis ($12.5$, $24.7$, $27.0$, $21.0$, and $13.9\ms{}$) and peaks at the three-layer head. The two ends narrow it for opposite reasons. A linear head is too weak to exploit either encoder. A full-attention head recovers much of the sound speed from random features on its own ($36.22\ms{}$, doing its own multi-angle fusion), so the representation adds less there. In the middle, a mid-capacity head on random features stays far worse than on the self-supervised features ($48.37$ versus $27.37\ms{}$ at the dilated head), where the representation matters most.

\begin{narrowfigure}[tb]
 \centering
 \begingroup
 \renewcommand{\tiny}{\fontsize{6.5}{7.8}\selectfont}%
 \renewcommand{\scriptsize}{\fontsize{9.1}{10.9}\selectfont}%
 \renewcommand{\footnotesize}{\fontsize{10.4}{12.5}\selectfont}%
 \renewcommand{\small}{\fontsize{11.7}{14}\selectfont}%
 \renewcommand{\normalsize}{\fontsize{13}{15.6}\selectfont}%
 \adjustbox{max width=\linewidth}{%
\begin{tikzpicture}
  \begin{axis}[
      iqaxis,
      width=9cm, height=7cm,
      symbolic x coords={linear,conv,conv3,dilated,attn},
      xtick=data,
      xlabel={probe head (increasing receptive field)},
      ylabel={sound-speed MAE (m/s)},
      ymin=0, ymax=80,
      enlarge x limits=0.12,
      legend style={font=\footnotesize, draw=black!25, fill=white,
          at={(0.5,-0.24)}, anchor=north, legend columns=1, cells={anchor=west}},
    ]
    \addplot[iq/refline, forget plot] coordinates {(linear,15.60) (attn,15.60)};
    \node[anchor=south west, iqamber, font=\scriptsize] at (axis cs:linear,15.60) {full SSL+ft = 15.60};

    \addplot[iq/ssl] coordinates {
        (linear,62.05) (conv,45.64) (conv3,36.74) (dilated,27.37) (attn,22.34)
      };
    \addlegendentry{SSL frozen}

    \addplot[iq/random] coordinates {
        (linear,74.53) (conv,70.29) (conv3,63.70) (dilated,48.37) (attn,36.22)
      };
    \addlegendentry{random frozen}
  \end{axis}
\end{tikzpicture}}%
 \endgroup

 \caption{Frozen-probe head-capacity ladder on the self-supervised encoder versus a randomly initialized encoder, the decoder-capacity control. Values are sound-speed test MAE at 10{,}000 labels, with the fine-tuned model as reference. The self-supervised features beat random at every head capacity, and the advantage is an inverted-U peaking at intermediate capacity, so the readout reflects the representation rather than the probe head.}
 \label{fig:appendix-probe-ladder}
\end{narrowfigure}

\section{The $U(1)$ symmetry by construction}
\label{sec:appendix-symmetry}

With no training and no labels, we rotate the global input phase $z \mapsto e^{i\psi} z$ and check how each component $f$ responds. The conjugate-product feed-forward should be invariant, $f(e^{i\psi} z) = f(z)$. It is (relative change 0.000, against 0.959 for a pointwise feed-forward). The Hermitian attention should be equivariant, $f(e^{i\psi} z) = e^{i\psi} f(z)$. It is (deviation 0.005 from the expected rotation, against 1.193 for an unconstrained real attention that scrambles the phase). The components therefore carry the demodulation symmetry by construction. The complex patch embedding is the equivariant front end that gates both. On a real patch embedding, the signatures collapse.

\section{Fine-tuning fuses the channel data into the spatial domain}
\label{sec:appendix-finetune-projection}

\begin{figure}[tb]
 \centering
 \includegraphics[width=0.7\linewidth]{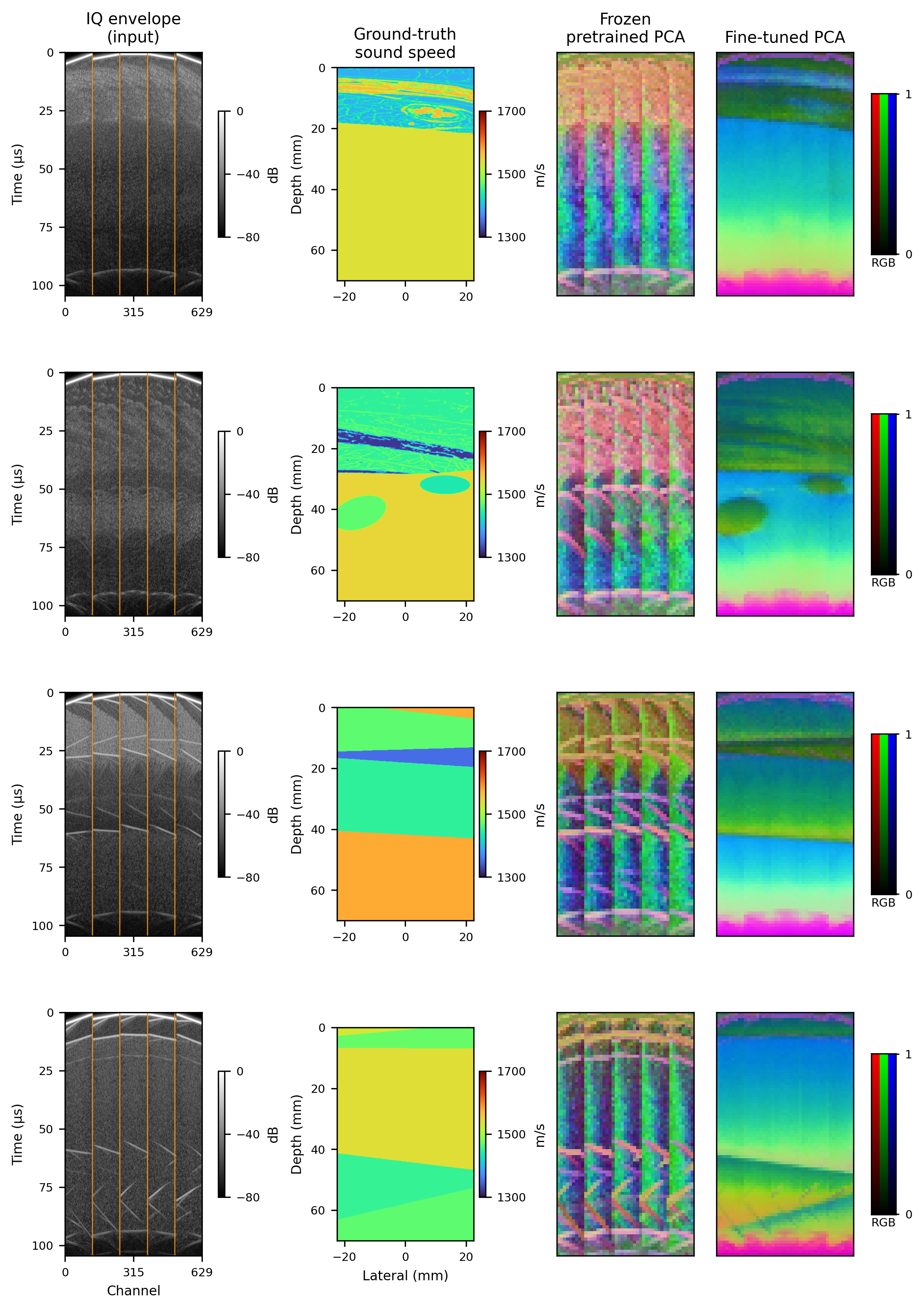}
 \caption{Frozen pretrained versus fine-tuned features, for four held-out samples (rows). Columns are the IQ envelope input, the ground-truth sound speed, the frozen JEPA-pretrained features, and the same encoder's features after fine-tuning, with the last two projected to RGB by their top three principal components (independent basis per column). Fine-tuning removes the vertical angle-concatenation seams and reorganizes the features into the spatial sound-speed layout.}
 \label{fig:finetune-projection}
\end{figure}

The same principal-component read-out, applied before and after fine-tuning, shows what fine-tuning changes in the representation (Figure~\ref{fig:finetune-projection}). The frozen pretrained features still carry the acquisition layout, with vertical seams visible between the concatenated plane-wave angles. The vertical striping of the complex encoder (Appendix~\ref{sec:appendix-latent-architecture}) is visible here too, in all four samples. After fine-tuning on sound speed those seams are gone. The features are reorganized into the spatial layout of the sound-speed map, with depth-aligned layers and inclusions. The model has learned to fuse the multi-angle channel data and map it into the image domain internally. The supervised objective only refines this implicit compounding and migration rather than building it from scratch.

\section{Extended qualitative predictions across phantom families}
\label{sec:appendix-gallery}

\begin{figure}[tb]
 \centering
 \includegraphics[width=1.0\linewidth]{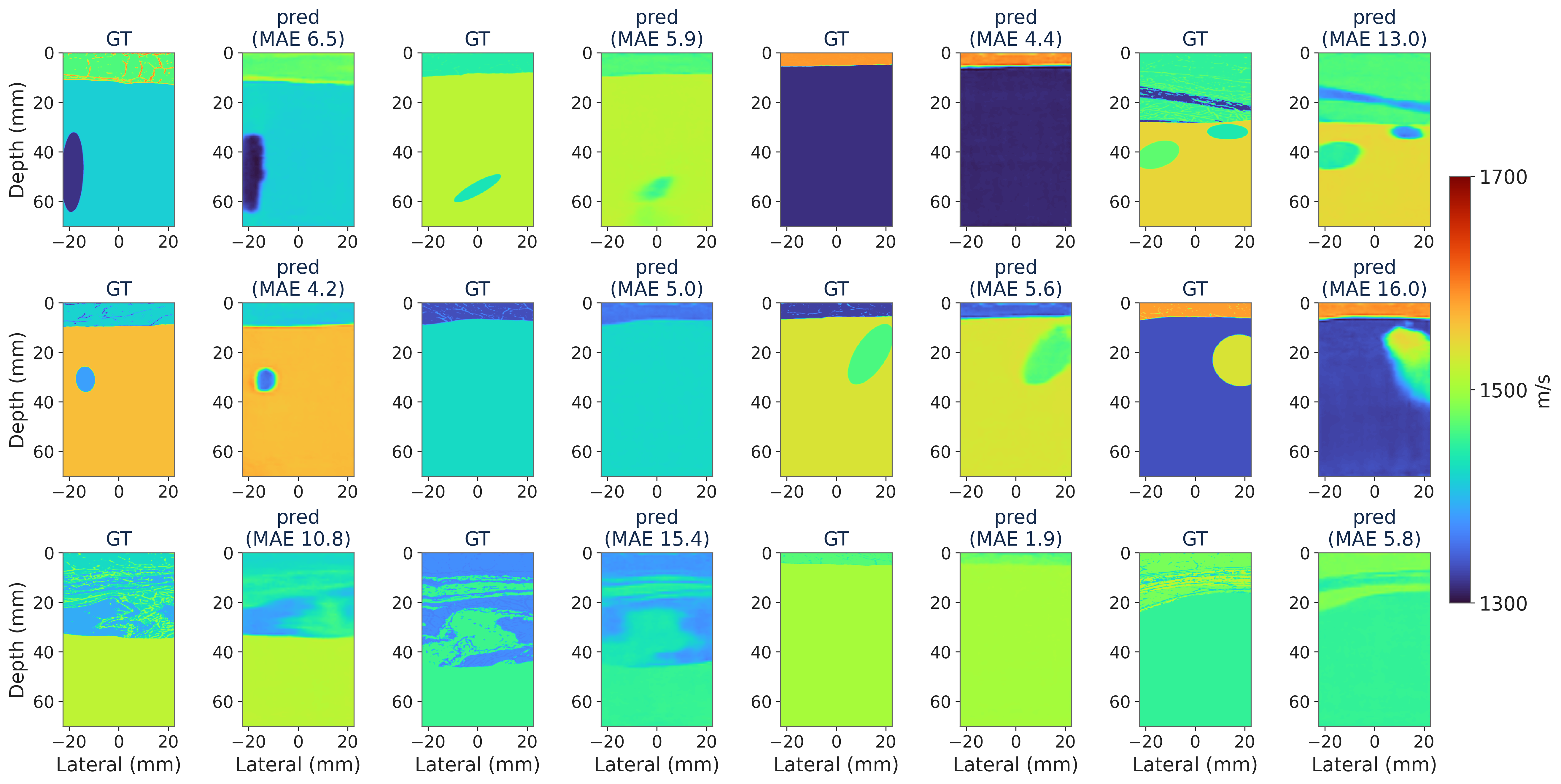}\\[1.5ex]
 \includegraphics[width=1.0\linewidth]{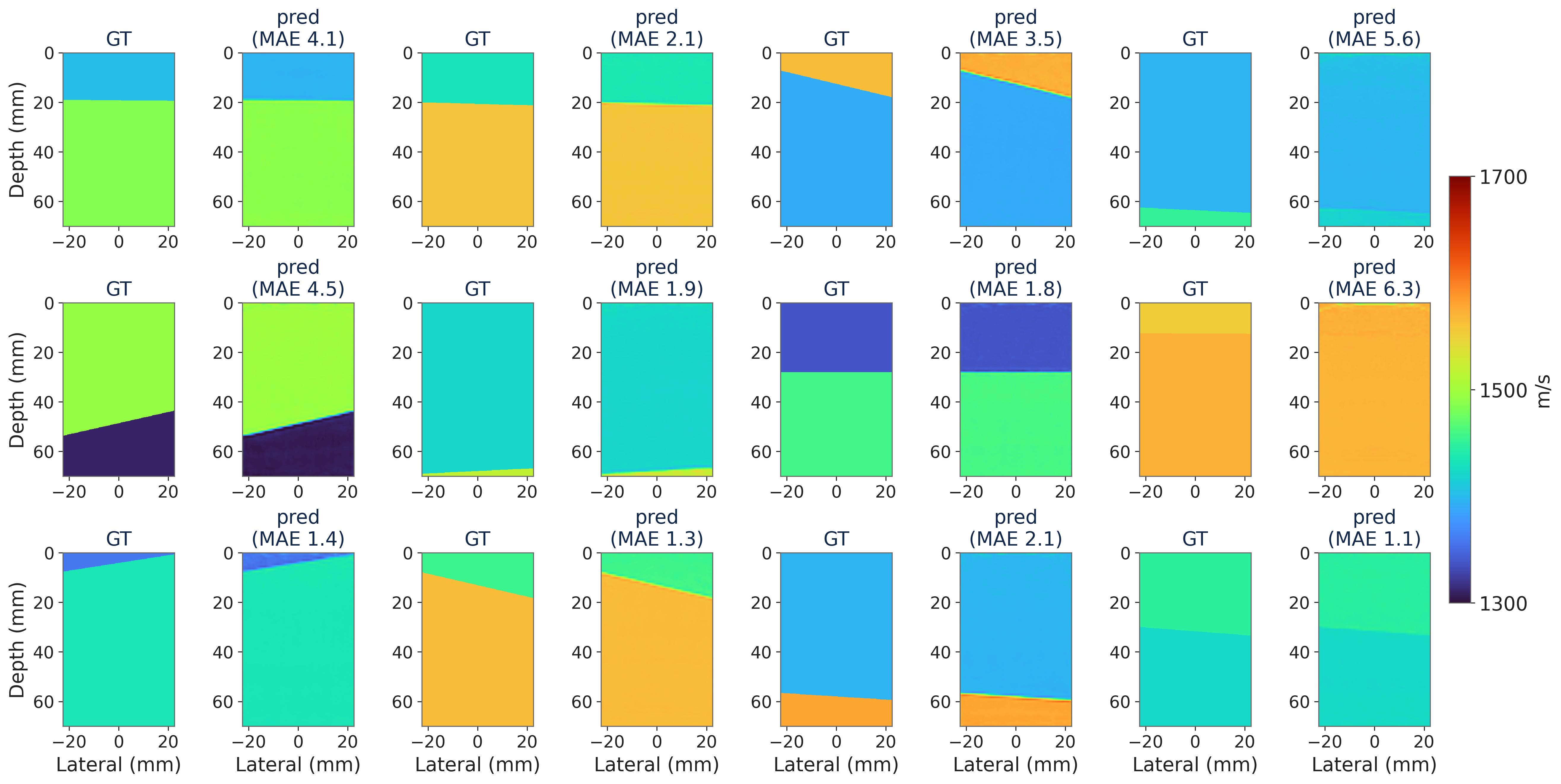}
 \caption{Held-out prediction galleries for the IQ-JEPA Hermitian ViT at 63{,}435 labels, abdominal (top) and random 2-layer (bottom) families. Each panel is ground truth versus prediction with per-sample MAE.}
 \label{fig:appendix-gallery-1}
\end{figure}

\begin{figure}[tb]
 \centering
 \includegraphics[width=1.0\linewidth]{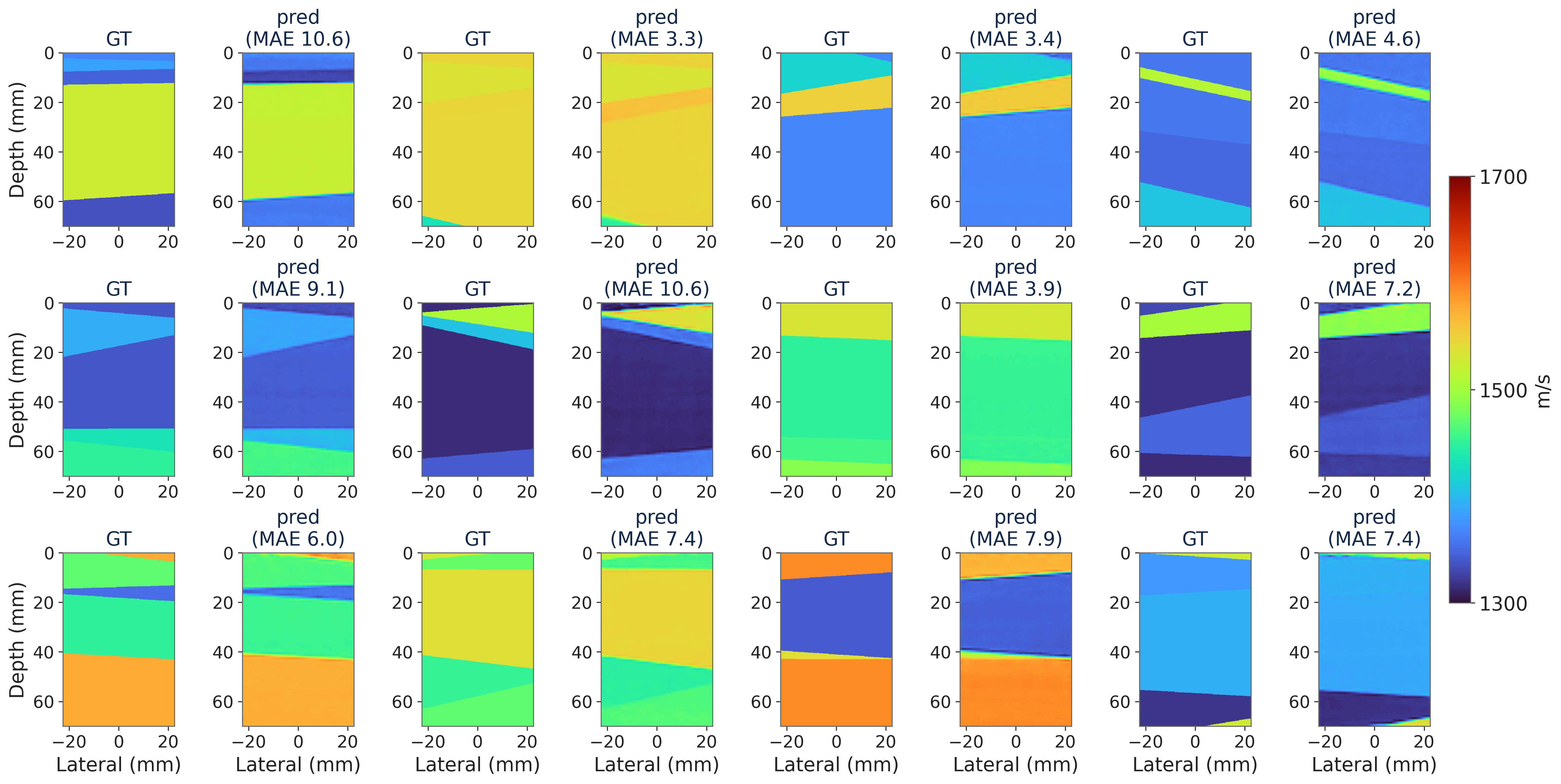}\\[1.5ex]
 \includegraphics[width=1.0\linewidth]{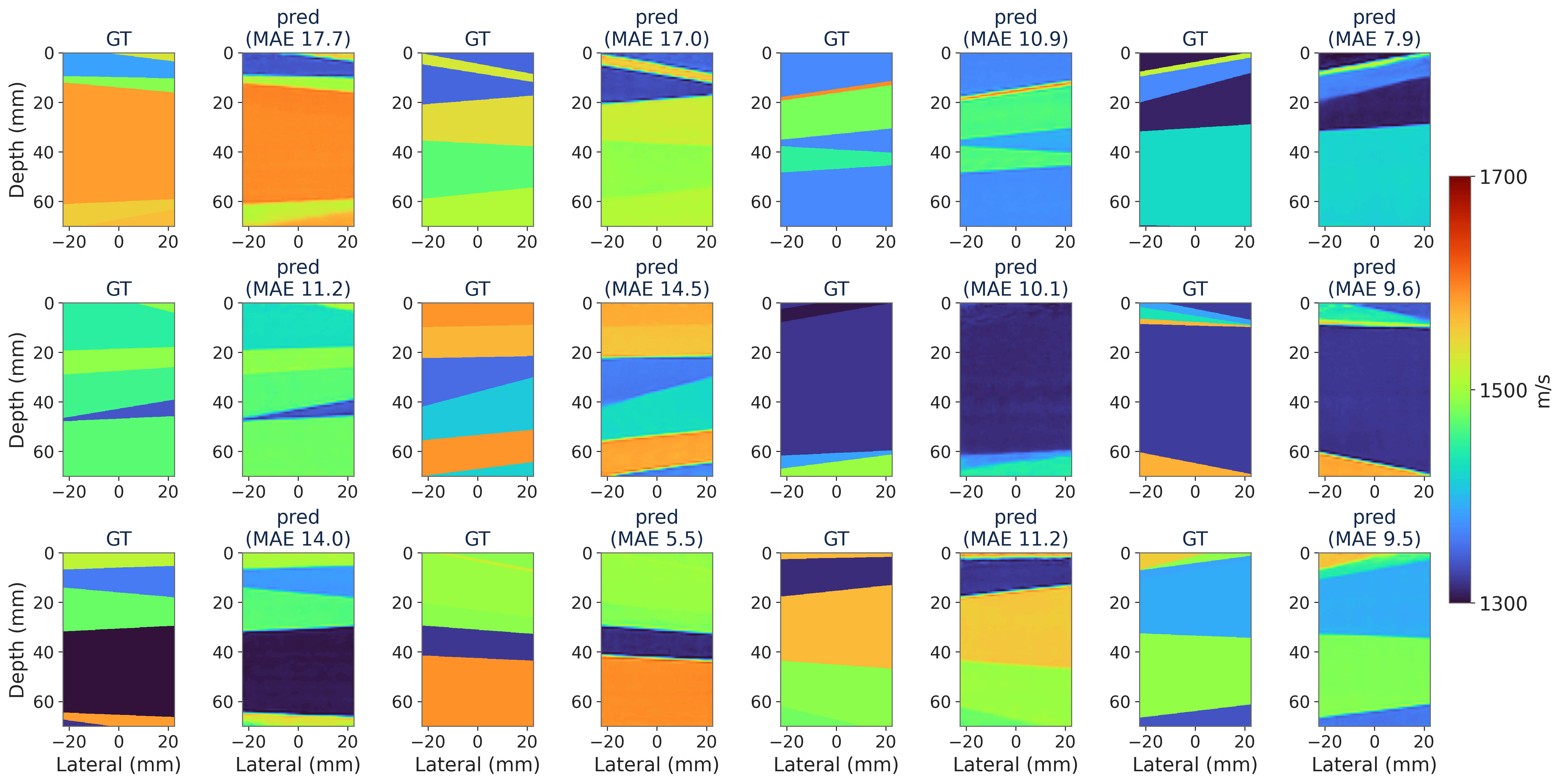}
 \caption{Held-out prediction galleries for the IQ-JEPA Hermitian ViT at 63{,}435 labels, random 5-layer (top) and 6-layer (bottom) families. Each panel is ground truth versus prediction with per-sample MAE.}
 \label{fig:appendix-gallery-2}
\end{figure}

The qualitative comparison in Figure~\ref{fig:sos-predictions} shows two representative samples. Figures~\ref{fig:appendix-gallery-1} and~\ref{fig:appendix-gallery-2} extend it to galleries of held-out samples that span the abdominal and random-layer families. Each panel shows the ground-truth sound speed against the IQ-JEPA prediction at 63{,}435 labels with per-sample MAE. The predictions recover the layer boundaries, inclusions, and speed ordering across phantom types, consistent with the per-family examples in Figure~\ref{fig:sos-predictions}.

\section{Extended inter-model comparison}
\label{sec:appendix-intermodel}

\begin{figure}[tb]
 \centering
 \includegraphics[width=0.7\linewidth]{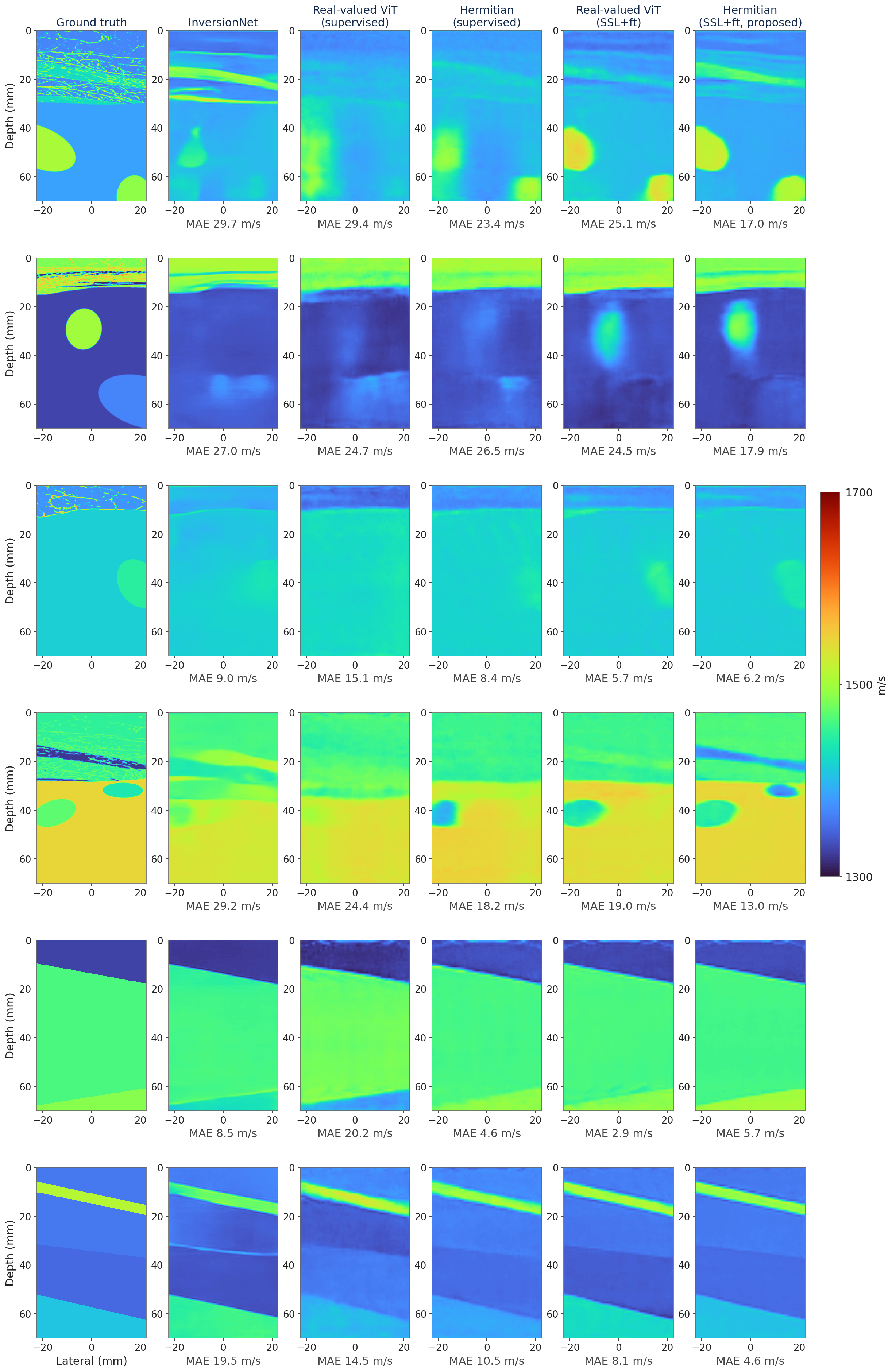}
 \caption{Extended inter-model sound-speed comparison at 63{,}435 labels on additional held-out samples (four abdominal, one random 3-layer, one random 5-layer). Columns are ground truth, InversionNet, the supervised real and Hermitian ViTs, the self-supervised real-valued ViT, and the IQ-JEPA Hermitian ViT, with per-sample MAE.}
 \label{fig:appendix-intermodel}
\end{figure}

The inter-model comparison in Figure~\ref{fig:model-comparison} shows two representative samples. Figure~\ref{fig:appendix-intermodel} extends it to further held-out samples spanning the abdominal and random-layer families. Each row shows the ground truth against InversionNet, the supervised real-valued ViT, the supervised Hermitian ViT, and the IQ-JEPA Hermitian ViT at 63{,}435 labels with per-sample MAE. The pattern of Figure~\ref{fig:model-comparison} holds across samples. The convolutional baseline smooths thin layers that the complex models retain.

\end{document}